\documentclass[twoside]{article}
\usepackage[accepted]{aistats_cameraready_package/aistats2023}

\usepackage{xcolor}
\usepackage{natbib}

\usepackage{booktabs}
\usepackage{enumitem}
\DeclareUnicodeCharacter{0010}{!!!} 

\usepackage{array}
\newcolumntype{L}[1]{>{\raggedright\let\newline\\\arraybackslash\hspace{0pt}}m{#1}}
\newcolumntype{C}[1]{>{\centering\let\newline\\\arraybackslash\hspace{0pt}}m{#1}}
\newcolumntype{R}[1]{>{\raggedleft\let\newline\\\arraybackslash\hspace{0pt}}m{#1}}

\usepackage[hyphens]{url}
\usepackage{hyperref}       % hyperlinks
\hypersetup{
     colorlinks,
     citecolor=blue,
     filecolor=blue,
     linkcolor=blue,
     urlcolor=blue,
}

\usepackage{amsmath,amssymb}

\usepackage{graphicx}
\usepackage{caption}
\begingroup
        \expandafter\ifx\csname pdfsuppresswarningpagegroup\endcsname\relax\else\global\pdfsuppresswarningpagegroup=1\relax\fi
\endgroup

\usepackage{algorithm}
\usepackage{algorithmic}

\usepackage{newfloat}
\usepackage{listings}
\DeclareCaptionStyle{ruled}{labelfont=normalfont,labelsep=colon,strut=off} % DO NOT CHANGE THIS
\floatstyle{ruled}
\newfloat{listing}{tb}{lst}{}
\floatname{listing}{Listing}

\begin{document}

%% TITLE AND AUTHORS
\twocolumn[
\aistatstitle{Fix-A-Step: Semi-supervised Learning From Uncurated Unlabeled Data}
% \aistatsauthor{ Author 1 \And Author 2 \And  Author 3 }
% \aistatsaddress{ Institution 1 \And  Institution 2 \And Institution 3 } ]
% \author {
% 	Zhe Huang\textsuperscript{\rm 1},
% 	Mary-Joy Sidhom\textsuperscript{\rm 1},
% 	Benjamin S. Wessler\textsuperscript{\rm 2},
%     Michael C. Hughes\textsuperscript{\rm 1}
% }
% \affiliations {
%     \textsuperscript{\rm 1} Dept. of Computer Science, Tufts University\\
%     \textsuperscript{\rm 2} Tufts Medical Center\\
% }
\aistatsauthor {
	Zhe Huang\textsuperscript{\rm 1},
	Mary-Joy Sidhom\textsuperscript{\rm 1},
	Benjamin S. Wessler\textsuperscript{\rm 2},
    Michael C. Hughes\textsuperscript{\rm 1}
}
\aistatsaddress {
    \textsuperscript{\rm 1} Dept. of Computer Science, Tufts University, Medford, MA, USA\\
    \textsuperscript{\rm 2} Division of Cardiology, Tufts Medical Center, Boston, MA, USA\\
}]
%% MCH SPACE-REDUCING HACK 2/3
% Reduce spacing around floats 
\setlength{\floatsep}{8pt plus 4pt minus 4pt}
\setlength{\textfloatsep}{8pt plus 4pt minus 3pt}

%% MCH SPACE-REDUCING HACK 3/3
% Reset vertical space for equations 
% (must be after \begin{document})
\setlength{\abovedisplayskip}{2pt plus 3pt}
\setlength{\belowdisplayskip}{2pt plus 3pt}

\begin{abstract}
Semi-supervised learning (SSL) promises improved accuracy compared to training classifiers on small labeled datasets by also training on many unlabeled images.
In real applications like medical imaging, unlabeled data will be collected for expediency and thus \emph{uncurated}: possibly different from the labeled set in classes or features.
Unfortunately, modern deep SSL often makes accuracy worse when given uncurated unlabeled data.
Recent complex remedies try to detect out-of-distribution unlabeled images and then discard or downweight them.
Instead, we introduce Fix-A-Step, a simpler procedure that views all uncurated unlabeled images as potentially helpful.
Our first insight is that even uncurated images can yield useful augmentations of labeled data.
Second, we modify gradient descent updates to prevent optimizing a multi-task SSL loss from hurting labeled-set accuracy.
Fix-A-Step can ``repair'' many common deep SSL methods, improving accuracy on CIFAR benchmarks across all tested methods and levels of artificial class mismatch.
On a new medical SSL benchmark called Heart2Heart, Fix-A-Step can learn from 353,500 truly uncurated ultrasound images to deliver gains that generalize across hospitals.

\end{abstract}

\section{INTRODUCTION}
A key roadblock to applying supervised learning to real applications is the need to assemble a large-enough labeled dataset for the intended task.
Modern deep learning pipelines are especially data-hungry.
In many cases, the acquisition of a large dataset of \emph{unlabeled} features is rather affordable.
However, providing reliable labels for each example is cost-prohibitive, often requiring expensive, time-consuming work from human experts.
This tradeoff is especially apt in our motivating application: classifying medical images where
images are collected in the course of routine care and easily available by querying a hospital's electronic records.
However, labeling images often requires clinical staff with years of training to spend minutes per image.

If only a tiny labeled set is available but we can access a big \emph{unlabeled} set of images, one promising approach is \emph{semi-supervised learning} (SSL)~\citep{zhuSemiSupervisedLearningLiterature2005,vanengelenSurveySemisupervisedLearning2020}. 
Recent years have seen amazing progress on standard benchmarks such as recognizing address digits from photos of houses (SVHN;~\citet{netzerReadingDigitsNatural2011}).
With only 100 labeled examples per digit class, a supervised neural net's error rate is roughly 12\%. 
Using a large unlabeled set, the FixMatch SSL method~\citep{sohnFixMatchSimplifyingSemisupervised2020} delivers error below 2.5\%, while even more recent work has pushed below 2\%~\citep{xuDPSSLRobustSemisupervised2021, hanUnsupervisedSemanticAggregation2020}.

Unfortunately, common benchmarks like SVHN may be too optimistic. 
In real tasks, unlabeled sets will be collected automatically at scale for convenience, and thus \emph{uncurated}: they may differ from the labeled set in terms of represented classes, class frequencies, or even features.
% To maintain the efficiency that motivates SSL, the effort required to curate the unlabeled set -- that is, apply labels and assess differences -- remains prohibitive.
Effective SSL must improve accuracy despite such uncurated data.

Off-the-shelf SSL using mismatched unlabeled sets often predicts \emph{worse} than just ignoring unlabeled data~\citep{oliverRealisticEvaluationDeep2018,calderon-ramirezDealingScarceLabelled2021}.
Recent methods try to be robust to unlabeled sets that differ from the labeled set (see Tab.~\ref{tab:related_work_comparison}).
% line of work called ``safe'' or ``open-set'' SSL.
%deliver accuracy that is at least as good as using the labeled set alone but better when possible.
The dominant paradigm is intuitive: 
identify examples in the unlabeled set that are \emph{out-of-distribution} (OOD), then remove or downweight them \citep{calderon2022semi,chen2022semi, he2022not}.
We find this line of work delivers insufficient gains in accuracy, while adding complexity due to OOD detection and discarding a substantial amount of unlabeled data.
%The resulting algorithms thus depend heavily on the OOD detector's performance.
% which adds additional complexity \todo{while delivering only modest gains.} 

% Larger, more reliable gains are needed to enable SSL in real uncurated applications.
%\todo{Further, viewing OOD samples as completely useless limited the possible performance of SSL algorithms in real applications, since a substantial portion of the unlabeled data could be discarded}. 

This study makes 3 contributions toward robust SSL backed by reproducible experiments\footnote{Code and Heart2Heart data: \href{https://github.com/tufts-ml/fix-a-step}{github.com/tufts-ml/fix-a-step}}.
First, we challenge the dominant paradigm of filtering out or downweighting OOD examples in the unlabeled set.
Our experiments suggest that even perfect OOD filtering, which is unrealistic in practice, does not perform well (see Fig.~\ref{fig:results-cifar10}).
Instead of viewing OOD images as probably harmful, we argue for a \textbf{new paradigm: OOD images from uncurated unlabeled sets are possibly helpful}.
%this sentence need to break down?

Second, following this paradigm we introduce \textbf{a new training procedure called \emph{Fix-A-Step} that delivers accuracy gains from uncurated unlabeled sets}.
When applied to repair several deep SSL methods across a range of labeled-unlabeled class mismatch levels, our Fix-A-Step improves predictions better than alternative methods while being substantially simpler and faster too.

Finally, we offer a \textbf{new SSL benchmark called Heart2Heart that uses truly uncurated unlabeled medical images and assesses cross-hospital generalization}.
%For too long, SSL methods advances have been evaluated only on the same repurposed generic object recognition datasets. 
Using three inter-operable open-access datasets -- TMED~\citep{huangNewSemisupervisedLearning2021}, CAMUS~\citep{leclercDeepLearningSegmentation2019}, and Unity~\citep{howardAutomatedLeftVentricular2021} --
we pursue a clinically-relevant problem: recognizing the view type of an echocardiogram image of the heart.
Future methods that learn from limited data can follow our reproducible protocol. 
We hope this new Heart2Heart benchmark enables authentic SSL applications in medicine and ultimately improves care for patients with heart disease.

\begin{figure}
\includegraphics[width=0.47\textwidth]{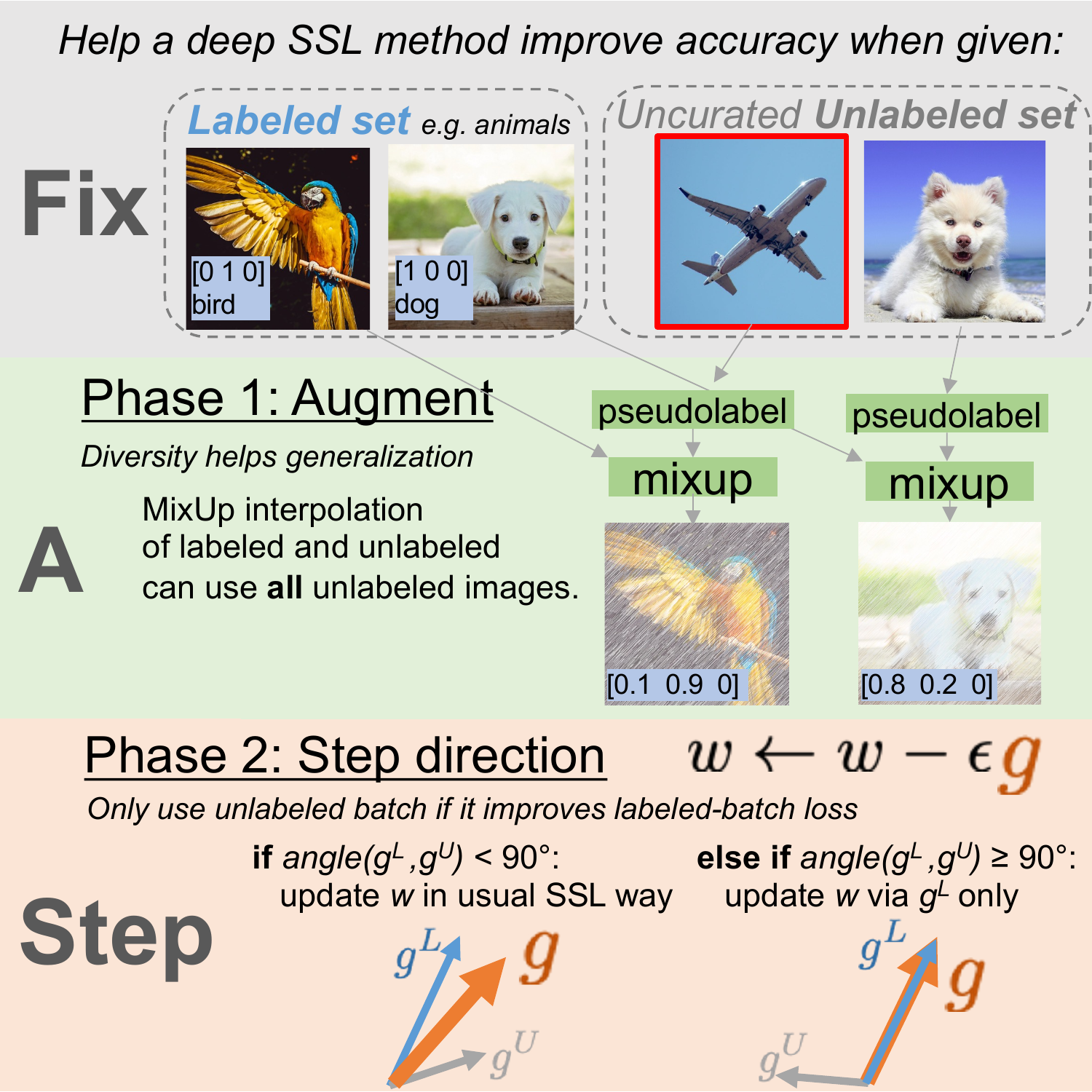}
\caption{
Diagram of our Fix-A-Step approach, which can be used to improve accuracy despite \emph{uncurated data} for any SSL method (e.g. VAT or FixMatch) that trains via a loss like Eq.~\eqref{eq:loss-for-consistency-ssl}.
}%endcaption	
\label{fig:fixastep_diagram}
\end{figure}

\section{BACKGROUND \& RELATED WORK}
\label{sec:background}
\label{sec:related_work}

We pursue \emph{semi-supervised} learning (SSL)~\citep{zhuSemiSupervisedLearningLiterature2005,vanengelenSurveySemisupervisedLearning2020} for the specific problem of image classification with deep neural networks~\citep{oliverRealisticEvaluationDeep2018}.
We can observe images $x$ (represented as $D$-dimensional vectors) as well as corresponding labels $y \in \{1, 2, \ldots C\}$ for $C$ classes of interest.
The goal is to train predictors from both a labeled dataset $\mathcal{D}^L$ of feature-labeled pairs $x,y$ and an unlabeled dataset $\mathcal{D}^U$ containing feature vectors $x$ only.

%% MCH: Cut this bc too redundant with earlier
%In our motivating applications, unlabeled examples are easily collected, while labels are difficult due to burdensome requirements of time, money, or human expertise.
%Medical image classification is well-suited to SSL, because routinely collected images are available in abundance, but annotation costs usually mean only small labeled sets are affordable. 

%%%%%% TABLE: RELATED WORK
\setlength{\tabcolsep}{.24cm}
\begin{table*}[!t]
\begin{tabular}{l r  l l l}
Method & Acc. & Paradigm 
	%& \todo{WHAT ABOUT DELETING THIS COLUMNApplicability} 
	& Extra Complexity & Realistic Eval.
\\ \hline
Fix-A-Step (ours) 
	& 85.4 % 85.45 verified in google doc, MT_CombinedAugGrad
	& OOD \emph{helpful}
	%& any $\ell^U$
	& none
	& Heart2Heart
\\
TOOR {\tiny \citep{huangTheyAreNot2022}}
	& $*$78.5 % read from Fig 4 of Huang et al TOOR paper
	& OOD \emph{helpful}
	%& any $\ell^U$
	& Separate NN for OOD discrimination
	& none
\\
CL {\tiny{\citep{cascante-bonillaCurriculumLabelingRevisiting2021}}}
	& $*$83.0
	& OOD harmful % Quote from paper: "We conjecture that our self-pacing curriculum is key to this scenario, where the adaptive thresholding scheme could help filter the out-of-distribution unlabeled samples during training"
	%& pseudo-label $\ell^U$
	& Multiple rounds of training, each from scratch
	& none
\\
OpenMatch {\tiny{\citep{saitoOpenMatchOpensetConsistency2021}}}
	& 82.3 % 82.25
	& OOD harmful
	%& \todo{FixMatch only?}
	& Extra one-vs-all OOD detector / class
	& none
\\
DS3L {\tiny \citep{guoSafeDeepSemiSupervised2020}}
	& 74.7 %74.7 from our implementation 
	% in google doc, Row "DS3L + MT (reported)"
	% sanity check: looks like 77.5 in their Fig 6
	& OOD harmful
	%& any $\ell^U$
	& 3x train time due to bilevel optimization
	& none
\\
MTCF {\tiny \citep{yuMultiTaskCurriculumFramework2020}}
	& 77.0 % read from our Fig 1 (after adding cosine LR, 68.0-->77.0)
	& OOD harmful
	%& any $\ell^U$
	& Extra OOD head, curriculum learning
	& none
\\
UASD {\tiny \citep{chenSemiSupervisedLearningClass2020}}
	& $*$78.2 % in google doc, row "UASD Figure 3 left (reported)"
	& OOD harmful
	%& fixed $\ell^U$
	& none
	& none
\\
Safe-Student
	{\tiny{\citep{heSafeStudentSafeDeep2022}}}
	& $^{\dagger}$n/a
	& OOD harmful
	%& \todo{FILL} %their objective eq11 doesnt fit our eq1?
	& 2 NNs (teacher \& student), extra KL loss
	& none
\end{tabular}
\caption{Comparison of related work on open-set/safe SSL.
\emph{Acc} means accuracy on the CIFAR-10 6-animal task (defined in Sec.~\ref{sec:results-cifar}) with 400 labeled examples/class and an open-set unlabeled set (50\% mismatch).
Fix-A-Step uses a FixMatch base model, as does OpenMatch.
%, or MeanTeacher for older methods.
Numbers come from our implementation except if marked \textbf{*} (copied from cited paper) or $^\dagger$ (not assessed in cited paper).
%\emph{Time:} Runtime for training on 6-animal task.
\emph{Paradigm:} how each method treats out-of-distribution (OOD) images in the unlabeled set, broadly either possibly helpful or likely harmful (and thus in need of filtering).
%\emph{Applicability:} whether the method might be possible with different unlabeled losses $\ell^U$.
\emph{Extra Complexity:} additional neural networks, layers, or runtime concerns that exceed a standard SSL deep classifier like MixMatch.
\emph{Realistic Eval.:} evaluation beyond ``artificial'' unlabeled sets from common datasets like CIFAR, ImageNet, etc.
%$\dagger$: \todo{DELETE? OpenMatch has no realistic SSL evaluation, but evaluates OOD detection on fine-grained data (e.g. breeds of dog).}
%\todo{CITE} and Dogs\todo{CITE}.
}%endcaption	
\vspace{-0.1cm}
\label{tab:related_work_comparison}
\end{table*}

\paragraph{Training for Deep SSL.}
While many SSL paradigms have been tried~\citep{kingmaSemisupervisedLearningDeep2014,kumarSemisupervisedLearningGANs2017,nalisnickHybridModelsDeep2019}, the dominant approaches for semi-supervised training of deep image classifiers today continue to modify standard objectives for discriminative neural nets by adding a regularization term using unlabeled data~\citep{miyatoVirtualAdversarialTraining2019,sohnFixMatchSimplifyingSemisupervised2020}.
This approach trains a neural net probabilistic classifier $f$ with weights $w$ by solving the optimization problem:
\begin{align} % Use \! to reduce whitespace between symbols
\min_{w} ~ \!
	\textstyle \sum_{x,y \in \mathcal{D}^L} \!
		\ell^L( y, f_w(x) )	
	+ 
	\lambda \!
	\sum_{x \in \mathcal{D}^U} \!
		\ell^U(x; w)
	\label{eq:loss-for-consistency-ssl}
\end{align}
Here, the first term $\ell^L$ is a labeled-set-only \emph{cross entropy} loss and the second term $\ell^U$ is a method-specific unlabeled-set loss.
A key hyperparameter is the unlabeled-loss-weight $\lambda > 0$, which balances the two terms.
Approaches such as the Pi-model~\citep{laineTemporalEnsemblingSemiSupervised2017}, Pseudo-Label~\citep{leePseudoLabelSimpleEfficient2013}, Mean-Teacher~\citep{tarvainenMeanTeachersAre2017}, Virtual Adversarial training (VAT)~\citep{miyatoVirtualAdversarialTraining2019}, and FixMatch~\citep{sohnFixMatchSimplifyingSemisupervised2020} all fit this objective, with variations in (1) the choice of function for $\ell^U$; (2) how data augmentation may alter images $x$; (3) procedures within $\ell^U$ that produce a perturbed image $x'$ that should be consistent with $x$; and (4) optimization routines to solve for $w$.

\textbf{Uncurated SSL.}
Unlabeled sets collected automatically at scale are by construction \emph{uncurated}, meaning their contents (features and true labels) are intended to be similar to the target labeled set but not carefully validated.
When the unlabeled set contains images from classes other than the $C$ classes represented in the labeled set, others call this ``open-set'' SSL~\citep{yuMultiTaskCurriculumFramework2020}. 
More formally, if we were to apply labels to the unlabeled set $\mathcal{D}^U$, the set of such labels may include an unknown number of classes beyond the $C$ known classes in the labeled set, and in the worst case may not even include any examples from some (or all) known classes.
Open-set SSL is a special case of uncurated SSL, because a truly uncurated dataset may also differ (usually slightly) in \emph{feature distributions} from the labeled set.
Our CIFAR evaluation focuses on open-set SSL, our Heart2Heart unlabeled set (Sec.~\ref{sec:results-tmed}) is truly uncurated.
% from medical applications.

\citet{oliverRealisticEvaluationDeep2018}~designed seminal experiments using CIFAR-10 images that purposefully build an unlabeled set (some animals, some not) that is mismatched in class composition from the labeled set (all animals).
As mismatch increases, many SSL methods (e.g. VAT or Pi-Model) score worse than a labeled-only baseline that ignores the unlabeled set.
Our later experiments confirm this (Fig.~\ref{fig:results-cifar10} left).

Several approaches have tried to remedy this deterioration, striving to be robust to open-set unlabeled data.
Such methods are also called \textit{safe SSL}~\citep{guoSafeDeepSemiSupervised2020}, because their goal is to perform no worse than labeled-set-only methods.
%UASD~\citep{chenSemiSupervisedLearningClass2020} filters out unlabeled examples which appear to be ``out of distribution'' according to confidence signals derived from its probabilistic classifier. 
%Alternatively, DS3L~\citep{guoSafeDeepSemiSupervised2020} learns a weighting function that can downweight unlabeled examples to prevent poor labeled-set performance.
Tab.~\ref{tab:related_work_comparison} summarizes previous works, with further discussion below.
These methods have been evaluated primarily on artificially mismatched remixes of datasets like CIFAR, and not yet on uncurated medical data.
% from potential SSL application domains like medicine. 

% Broadly, these existing safe SSL approaches do not perform as well as needed (see Fig.~\ref{fig:results-cifar10} far right).
%\todo{ DELETE? shows that the recent non-safe FixMatch dominates 3 ``safe'' SSL methods (UASD, DS3L, and MTCF).
% we remedy in our later medical evaluations.

%We now situate our work in the broader literature of methods focused on open-set SSL (also called safe SSL).
%In the supplement, we provide additional discussion of two other lines of related work: gradient step direction modifications and SSL for medical imaging.

\textbf{Related work: Open-set SSL that filters out OOD.}
%Recently, there are many efforts being made to tackle this real-world challenge. 
Most previous work on open-set SSL focuses on detecting then removing or downweighting OOD samples, assuming these harm the ultimate accuracy of an SSL classifier \citep{calderon2022semi,chen2022semi, he2022not}.
% under the assumption that OOD samples can only harm the ultimate accuracy of an SSL classifier 
\citet{chenSemiSupervisedLearningClass2020}'s UASD ensembles model predictions temporally to produce probability predictions for unlabeled samples, with confidence-based thresholding to filter out OOD samples. 
\citet{yuMultiTaskCurriculumFramework2020} propose a multi-task curriculum learning framework (MTCF) that alternates between updates to NN weights and updates to anomaly scores used to detect OOD images.
\citet{guoSafeDeepSemiSupervised2020}'s Deep Safe Semi-supervised Learning (DS3L) employs meta-learning ideas to downweight OOD samples. 
\citet{saitoOpenMatchOpensetConsistency2021}'s OpenMatch unifies FixMatch with novelty detection to learn representations of inliers while rejecting outliers.
\citet{heSafeStudentSafeDeep2022}'s Safe-Student use a teacher-student network that identifies OOD via an energy discrepancy score, while
\citet{baeSafeSemisupervisedLearning2022} filter OOD images via Bayesian neural networks.
%Although proposed under the open-set SSL setting, a main focus of the paper has been the effectiveness of the proposed OOD detection module. 
These methods have made notable strides on the class mismatch problem. 
However, they \emph{focus on reducing possible harm by filtering OOD images but neglect the potential benefits.}
% MCH: cut for space We will show later how OOD samples can be useful in SSL training.
% HZ says: So i changed the framing here a little bit: previously it sounds like two completely opposite position: OOD harmful vs OOD helpful. Now re-frame it as previous study only focus on one aspect of OOD: the possible harm. While we realize there is another aspect of OOD: possible benefit. ---> Therefore we try to maximize the benefit while alleviating the harm.

\textbf{Related work: Open-set SSL beyond filtering.}
Some recent work tries to \emph{detect} OOD images but still learn something useful from them.
For example, recent parallel work by
\citet{huangTheyAreNot2022} suggests OOD images may not be ``completely useless.''
Their TOOR method trains a model to classify in-distribution (ID) versus OOD images, and then, viewing OOD samples as from a related domain, pursue adversarial domain adaptation to ``recycle'' OOD samples.
\citet{luoEmpiricalStudyAnalysis2021} try to reduce the distribution gap between ID and OOD samples via style transfer:
transformed OOD samples are used as if they were ID samples in a consistency regularizer.
\citet{banitalebi2022auxmix}~detect ID and OOD samples, then use consistency regularization on ID samples and entropy maximization on OOD samples. 
\citet{huangTrashTreasureHarvesting2021}'s pretraining stage uses \emph{all unlabeled samples}, yet still filters out OOD samples later, assuming they would harm classifier accuracy.
\citet{cascante-bonillaCurriculumLabelingRevisiting2021} propose a curriculum labeling (CL) approach to SSL.
Over several training rounds, they increase the number of unlabeled images contributing pseudo-labels, eventually using all images.
Yet they conjecture their success is due to an
adaptive thresholding scheme  that can ``filter the out-of-distribution unlabeled samples''.
In contrast, our work does not need any OOD detector or filter at all; we treat all unlabeled images equally.

%% MCH TODO DISCUSS THIS WITH HZ
%\citep{brennanProviderlevelVariabilityTreatment2019} 
%\todo{DOES THIS BELONG HERE? Discuss \citet{laiSmoothedAdaptiveWeighting2022} }

%% MCH: We should keep these, helps reader appreciate broader context
\textbf{Other distantly related work.} 
\citet{ren2020not} learn a unique weight for each unlabeled sample for closed-set SSL. \citet{huangUniversalSemiSupervisedLearning2021} focus on cases where both class and feature distributions are mismatched.
\citet{caoOpenworldSemisupervisedLearning2022} study transductive learning for SSL in ``open worlds'' where novel classes appear in the unlabeled \emph{test set}.
%lthough the method is proposed under closed-set SSL, it seems can be extended to open-set SSL straightforwardly.}

% MCH: Maybe cite this in appendix?
%\todo{\citep{chen2019distributionally} propose robust SSL algorithm that reduce person-specific discrepancy and preserve task-specific consistency for people-centric sensing tasks.} \todo{\citet{killamsetty2021retrieve} use coreset selection to enable faster training, while achieving robustness properties.}  \todo{\citet{he2022not} partition the model parameters into ``safe'' and ``harmful'', they optimize the safe parameters while disabling the harmful parameters.}

%\textbf{Comparison to Fix-A-Step.}
%We compare our approach to others across multiple axes in Tab.~\ref{tab:related_work_comparison}.
%Unlike previous efforts, our Fix-A-Step approach is simpler both conceptually and in implementation cost: we do not require multiple stages or train any extra neural networks beyond a deep classifier.

\textbf{SSL benchmarks.}
SSL evaluations continue to focus on repurposed datasets such as CIFAR-10/100~\citep{krizhevskyLearningMultipleLayers2009}, or ImageNet~\citep{deng2009imagenet} (Tab.~\ref{tab:related_work_comparison}).
In App.~\ref{app:related_work}, we argue this exclusive focus is insufficient because (1) the SSL application is \emph{artificial}, dropping known labels to create unlabeled sets and (2) CIFAR specifically suffers from both label leakage due to perceptual duplicates~\citep{barzWeTrainTest2020} and incorrect labels~\citep{northcuttPervasiveLabelErrors2021}.
Some recent efforts strive to more realistically benchmark SSL algorithms~\citep{su2021realistic, wang2022usb}, but do not have a medical focus.
We hope our Heart2Heart benchmark and its truly uncurated unlabeled set
helps lead to impactful SSL for medical applications with plentiful images but hard-to-acquire labels.
% we hope our Heart2Heart benchmark with its realistically uncurated unlabeled set is a useful addition to the field.
%We argue that benchmarks using medical images with truly uncurated unlabeled sets are needed to properly assess SSL 

%It will likely include content that differs from the labeled set (new classes, different frequencies of classes, distribution shifts among features).

\textbf{Self-supervised learning.}
Another major way to learn from unlabeled data is self-supervised learning \citep{qi2020small}. Self-supervised learning aims at obtaining good feature representations or good network initialization without using manual annotations. Recent advance in self-supervised learning have achieved impressive results in closing the performance gap with supervised pre-training. \citep{chen2020simple, he2020momentum, chen2020improved}. While the goal is different, self-supervised learning could be adapted to semi-supervised learning setting, for example pre-training on all available data (labeled and unlabeled), and then fine-tuning on the labeled data \citep{caron2020unsupervised, chenBigSelfSupervisedModels2020}. However, \cite{saitoOpenMatchOpensetConsistency2021} reported that self-supervised learning does not help open-set SSL. We thus focus our experimental comparisons on semi-supervised methods here, and leave a more comprehensive investigation of self-supervision for future study.

\section{METHODS}
We have designed a training procedure for deep SSL classifiers that we call \emph{Fix-A-Step}, short for Fix via Augmentation and Step direction modification.
Fix-A-Step can be applied to any SSL method that minimizes an SSL objective matching Eq.~\eqref{eq:loss-for-consistency-ssl} via gradient descent.
Its goal is to make SSL classifiers robust to uncurated unlabeled data.

Fig.~\ref{fig:fixastep_diagram} illustrates the two key concepts of our approach.
First, unlabeled images, even when uncurated, can be \emph{helpful} in creating useful augmentations of the labeled set by injecting diversity.
Second, we protect against accuracy deterioration due to the unlabeled set by modifying the gradient update of neural net weights.
We do this not by filtering examples permanently, but by omitting the contribution of a batch's unlabeled loss gradient if its direction differs substantially from the labeled loss gradient. 
These two ideas are implemented in consecutive phases that occur when visiting each minibatch during gradient descent training.
Alg.~\ref{alg:FixAStep} provides pseudocode for Fix-A-Step, with further details below.
% algorithm is provided in Alg.~\ref{alg:FixAStep}.

\textbf{Phase 1: Augmentation.} In the \emph{Augmentation phase} (lines 3-4), our insight is to use \emph{all} unlabeled images in MixMatch-style augmentation~\citep{berthelotMixMatchHolisticApproach2019} of the labeled set.
We transform each labeled pair $(x,y)$ using another pair $(x', y')$ drawn either from the labeled set or the unlabeled set.
If only $x'$ is known, we use soft pseudo-label predictions for $y'$, see Alg.~\ref{alg:aug_softlabel}. 
Given $x, y$ and $x', y'$, we build a new labeled pair $\tilde{x}, \tilde{y}$ via MixUp~\citep{zhangMixupEmpiricalRisk2017} (see Alg.~\ref{alg:mixmatch}).
This new pair is used to compute the labeled loss. 
We readily acknowledge that the success of MixMatch for standard closed-set SSL is widely known.
However, for uncurated or open-set SSL, we believe MixMatch-style augmentation has been underexplored.

% is under-explored whether this technique is
%beneficial with uncurated data or compatible to other SSL algorithms. 

\begin{figure}[!b]
\centering
\begin{tabular}{c}
\includegraphics[width=.43\textwidth]{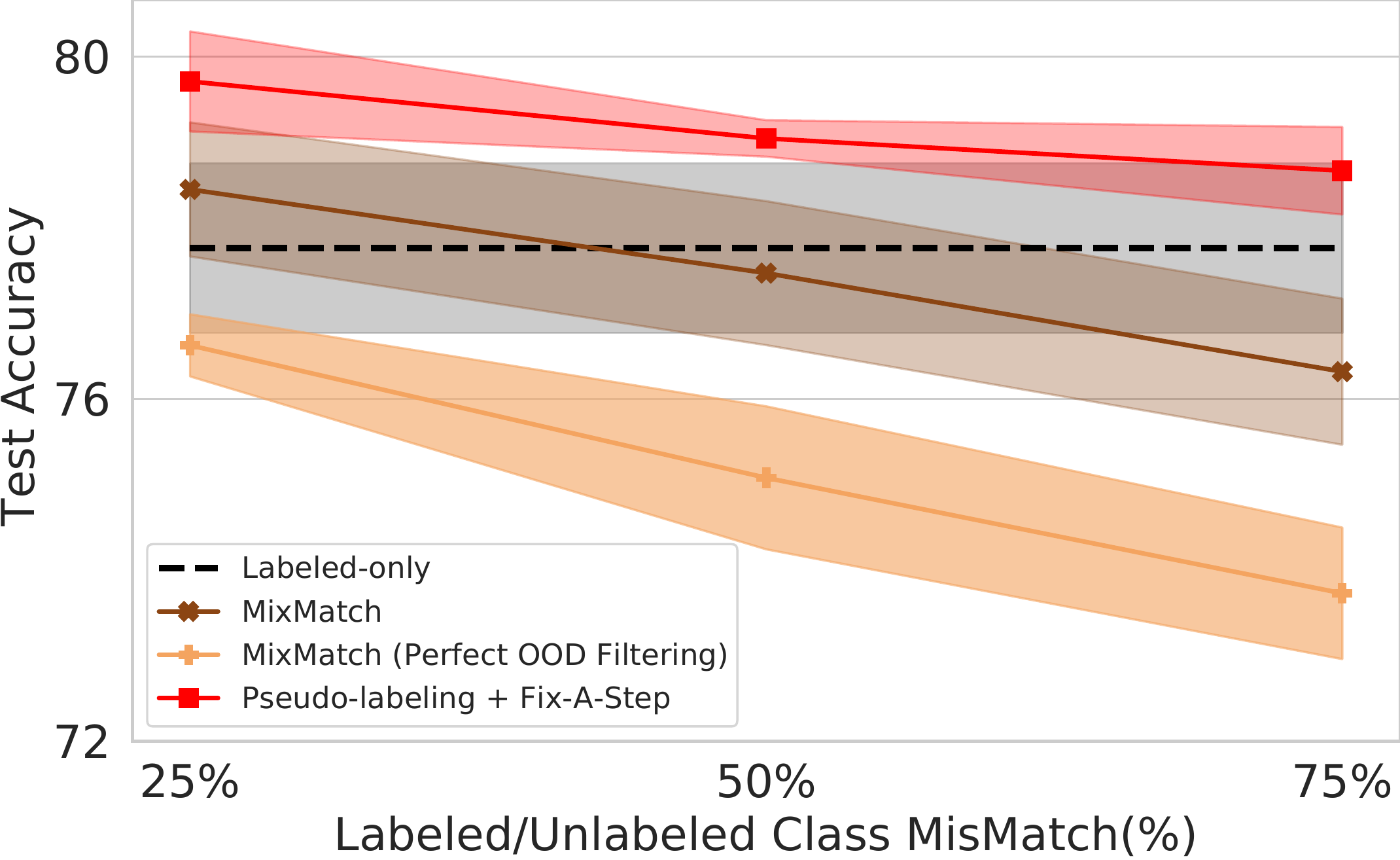}
\end{tabular}
\caption{
\textbf{Demo of benefits of Phase 1 (MixMatch using \emph{all} unlabeled images, even OOD) and Phase 2 (step direction)}, on CIFAR-10 6-animal task (400 examples/class).
Results average across 5 train/test splits (shaded area shows standard deviation).
%MixMatch with and without OOD sample cases use the same hyperparameters.
}%endcaption	
\label{fig:UnlabeledHelpfulMotivatingExample_MixMatch_main}
\end{figure}

Fig~\ref{fig:UnlabeledHelpfulMotivatingExample_MixMatch_main} shows the performance of standard MixMatch on the CIFAR-10 6-animal open-set task (Sec~\ref{sec:results-cifar}).
The first takeaway is that at each level of class mismatch, MixMatch \textbf{works better with OOD samples in the unlabeled set than without}. We call the latter ``perfect OOD filtering''. This suggests the value of augmenting with all unlabeled images.
% Details on the task description is in Sec ~\ref{sec:results-cifar}.}. We further show that MixMatch alone is not enough to handle contamination cases, see Fig ~\ref{fig:results-cifar10}
%draw interpolation weight $u \sim \text{Beta}(\alpha, \alpha)$, then construct $x^* \gets u x + (1-u) x'$, $y^* \gets u y + (1-u) y'$. 
%Unlabeled examples contribute diversity to augmentations.
The second takeaway is that \textbf{MixMatch-style augmentation alone is not enough}: beyond 25\% mismatch, the labeled-set only baseline matches or beats MixMatch.
Augmentation does not guard against the possible harm caused the unlabeled loss term, especially with OOD examples~\citep{saitoOpenMatchOpensetConsistency2021}. 

Fix-A-Step combines augmentation (phase 1) with protection against accuracy deterioration from the unlabeled loss (phase 2, described below) to ``repair'' SSL base models.
A final takeaway from Fig.~\ref{fig:UnlabeledHelpfulMotivatingExample_MixMatch_main} is that Fix-A-Step can repair a nine-year-old method, \citet{leePseudoLabelSimpleEfficient2013}'s Pseudo-label, to beat MixMatch across all mismatch levels.
While we tune hyperparameters for MixMatch here (see App.~\ref{app:hyperparameter_list}), to compare fairly \emph{we do not tune hyperparameters for Fix-A-Step}.

%%%%% ALGORITHM
\begin{algorithm}[!t]
\caption{Fix-A-Step Training}
\label{alg:FixAStep}
\textbf{Input}: Labeled set $\mathcal{D}^L$, Unlabeled set $\mathcal{D}^U$ (uncurated)
\\
\textbf{Output}: Trained weights $w$
\\
\textbf{Procedure} 
\linespread{1.1}\selectfont % improve spacing between lines
\begin{algorithmic}[1] %[1] enables line numbers
\FOR{iter $i \in 1, 2, \ldots I$ until converged}
\STATE $\mathbf{x}^L, \mathbf{y}^L, \mathbf{x}^U \gets {\small \textsc{GetNextMinibatch}}(\mathcal{D}^L, \mathcal{D}^U)$
\STATE $\tilde{\mathbf{x}}_1^U,\tilde{\mathbf{x}}_2^U,\tilde{\mathbf{y}}^U \gets {\small \textsc{Aug+PseudoLabel}}(\mathbf{x}^U; w, \tau)$
\STATE $\tilde{\mathbf{x}}^L, \tilde{\mathbf{y}}^L \gets {\small \textsc{MixMatchAug}}(\mathbf{x}^L, \mathbf{y}^L, \tilde{\mathbf{x}}_1^U, \tilde{\mathbf{x}}_2^U, \tilde{\mathbf{y}}^U; \alpha)$ 
\STATE $g^L \gets \nabla_w \ell^L( \tilde{\mathbf{x}}^L, \tilde{\mathbf{y}}^L; w)$
\STATE $g^U \gets \nabla_w \ell^U( \tilde{\mathbf{x}}_1^U, \tilde{\mathbf{y}}^U; w)$
%\IF {${\small \textsc{InnerProduct}}(g^L,g^U) > 0$}
%\STATE $w \gets w - \epsilon (g^L + \lambda_i g^U)$
%\ELSE
%\STATE $w \gets w - \epsilon g^L$
%\ENDIF
\STATE $w \gets \begin{cases}
w - \epsilon (g^L + \lambda_i g^U) ~~&\text{if~} \sum_d g^L_d g^U_d >0
\\
w - \epsilon g^L ~&\text{o.w.}
\end{cases}
$
\ENDFOR
\STATE \textbf{return} w
\end{algorithmic}
\textbf{Hyperparameters} (\emph{Values marked $\dagger$ tuned for all baselines as in App.~\ref{app:hyperparameter_list}. No tuning for Fix-A-Step in any experiment.})
\begin{itemize}[align=left,style=nextline,leftmargin=*,labelsep=3\parindent,font=\normalfont,topsep=0pt,itemsep=-1ex,partopsep=1ex,parsep=1ex]
	\item ~Temperature $\tau{=}0.5$ for {\small \textsc{Aug+PseudoLabel}} (Alg.~\ref{alg:aug_softlabel})
	\item ~Beta dist. shape $\alpha{=}0.5$ for {\small \textsc{MixMatchAug}} (Alg.~\ref{alg:mixmatch})
	\item ~Step size $\epsilon$$^\dagger$, Initial weights $w$, Max iterations $I$
	\item ~Unlabeled-loss weight per iter $\lambda_1, \ldots \lambda_I$$^\dagger$
\end{itemize}
\end{algorithm}

% \begin{algorithm}[tb]
% \caption{Fix-A-Step Training}
% \label{alg:FixAStep}
% \textbf{Input}: Labeled set $\mathcal{D}^L$, Unlabeled set $\mathcal{D}^U$ (uncurated)
% \\
% \textbf{Output}: Trained weights $w^*$
% \\
% \textbf{Hyperparameters}
% \begin{itemize}
% 	\item Shape parameter $\alpha{>}0$ of Beta dist. for {\small \textsc{MixMatchAug}}
% 	\item Initial weights $w$, Max. iterations $T$
% 	\item Unlabeled-loss weight per iter $\lambda_1, \ldots \lambda_T$
% \end{itemize}
% \begin{algorithmic}[1] %[1] enables line numbers
% \FOR{iter $t \in 1, 2, \ldots T$ until converged}
% \STATE $\{\mathbf{x}^L, \mathbf{y}^L\}, \mathbf{x}^U \gets {\small \textsc{GetNextMinibatch}}()$
% \STATE $\tilde{\mathbf{x}}^U, \tilde{\mathbf{y}}^U \gets {\small \textsc{Augment+SoftLabel}}(\mathbf{x}^U; w)$
% \STATE $\tilde{\mathbf{x}}^L, \tilde{\mathbf{y}}^L \gets 
% {\small \textsc{MixMatchAug}}(\{\mathbf{x}^L, \mathbf{y}^L\}, \tilde{\mathbf{x}}^U, \tilde{\mathbf{y}}^U; \alpha)$
% \STATE $g^L \gets \nabla_w \ell^L( \tilde{\mathbf{x}}^L, \tilde{\mathbf{y}}^L; w)$
% \STATE $g^U \gets \nabla_w \ell^U( \tilde{\mathbf{x}}^U, \tilde{\mathbf{y}}^U; w)$
% \IF {${\small \textsc{InnerProduct}}(g^L,g^U) > 0$}
% \STATE $w \gets w - \epsilon (g^L + \lambda_t g^U)$
% \ELSE
% \STATE $w \gets w - \epsilon g^L$
% \ENDIF
% \ENDFOR
% \STATE \textbf{return} w
% \end{algorithmic}
% \end{algorithm}

\textbf{Phase 2: Step direction modification.}
We address the possible harm from the unlabeled loss in phase 2 (lines 5-7 of Alg.~\ref{alg:FixAStep}), by modifying how neural net weights are updated.
The idea is to only use gradient information from the unlabeled loss if it improves labeled-set performance. 
%In lines 5-10 of Alg.~\ref{alg:FixAStep}, we prioritize the labeled loss in parameter updates, only using the unlabeled loss if it improves labeled-set performance. Compared to using complicated OOD detectors, our approach adds little complexity to exiting models}. 
%This way, we better ensure the diverse augmentations from phase one do not harm prediction quality.

At each batch, we compute two gradient vectors, one for each term in the loss: Let $g^L = \nabla_w \ell^L$ and let $g^U = \nabla_w \ell^U$.
%(using the MixMatch-derived $\tilde{x}, \tilde{y}$ from phase one),
Our Fix-A-Step update for weights $w$ using step size $\epsilon$ is
\begin{align}
w \gets \begin{cases}
 	w - \epsilon (g^L + \lambda g^U) \!\!\! &\text{if~} \sum_d g^L_d g^U_d > 0
 	\\
 	w - \epsilon g^L & \text{otherwise}.
 \end{cases}	
 \label{eq:weight-update-fixastep}
\end{align}
In the top case, we do the standard steepest descent update that minimizes the two-term SSL objective in Eq.~\eqref{eq:loss-for-consistency-ssl}.
In the bottom case, we perform an alternative update, using only the labeled-term gradient.
% \todo{From the transfer-interference trade-off perspective, the bottom case can be regarded as an extreme case of zero weight sharing between the labeled samples and unlabeled samples, when the learning of unlabeled samples interfere with the learning of labeled samples, while the first case is full weight sharing that tries to maximize the use of unlabeled data when the gradients are complementary.}
% that ensures we do not make labeled-set performance worse.
%the usual SSL update may not be a descent direction for the labeled loss $\ell^L$, and thus might
This two-case construction tries to ensure that SSL learning does not harm labeled set performance by ``turning off'' the gradient from an unlabeled batch when it interferes with the labeled loss.
We give geometric intuition below, then formally show that the gradient update in Eq.~\eqref{eq:weight-update-fixastep} always move weights $w$ in a \emph{descent direction} for the labeled set loss at the current minibatch.

\textbf{Geometric intuition for Phase 2.} Recall that two vectors $g^L$ and $g^U$ have positive inner product (top case update) only if the angle between the vectors is below 90 degrees, meaning their directions are similar.
At angles larger than 90 (bottom case), $g^L$ and $g^U$ are pointing in different directions, and minimizing the unlabeled loss would hinder the labeled loss.
In SSL, we care most about (heldout) classifier accuracy.
Any improvement on the unlabeled loss is useful only if it helps improve accuracy.
When $g^U$ points in a different direction than $g^L$, our update ignores the unlabeled gradient and updates weights $w$ using only $g^L$.
% if learning on unlabeled samples hinders learning of the labeled samples.}
% Thus, we can motivate the alternative lower case in Eq.~\eqref{eq:weight-update-fixastep} geometrically.
% %If the two gradients point in similar directions, we perform the usual SSL parameter update (top case).
% We usually care most about classifier accuracy, so when the two gradients point in opposing directions (bottom case), we might be better off ignoring the unlabeled and updating parameters using only the labeled loss.

\textbf{Definition 1: Descent direction of loss $\ell$.}
For any loss function $\ell$ for weight parameter vector $w \in \mathbb{R}^D$, a vector $v \in \mathbb{R}^D$ is a \emph{descent direction} of $\ell$ at $w$ if the inner product satisfies $v^T \nabla_w \ell < 0$ \citep{boydSecDescentMethods2004}.

\textbf{Lemma 1: The update in Eq.~\eqref{eq:weight-update-fixastep} steps in a descent direction of the labeled loss $\ell^L$ at the current minibatch.}
We prove for each case in Eq.~\eqref{eq:weight-update-fixastep}.
\emph{Top case}: By assumption, $\lambda > 0$ and the inner product $\sum_d g^L_d g^U_d$ is positive. This implies that $v{=}-(g^L + \lambda g^U)$ is a descent direction:
\begin{align}
v^T g^L =
	\underbrace{- \textstyle \sum_d (g^L_d)^2}_{\text{always negative}}
	\quad - \lambda \underbrace{\textstyle \sum_d g^L_d g^U_d}_{\text{pos. by assumption}}
< 0.
\end{align}
\emph{Bottom}: $-g^L$ is a descent direction for $\ell^L$ by definition.

While Lemma 1 provides a justification for our approach, we cannot guarantee the labeled loss will decrease after each step, for the same reasons that stochastic gradient descent (SGD) does not always decrease the loss after each update:
First, a descent direction of a minibatch may not be a descent direction of the entire dataset.
Second, step size matters; if $\epsilon > 0$  is too large, the loss may increase.
Nevertheless, with proper step size tuning, SGD has been wildly successful by following minibatch-specific descent directions.
Thus far, we find Fix-A-Step also successful.
% in practice.

\textbf{Inspiration from multi-task learning.}
Our step direction modification in Eq. (2) was developed independently but is similar to previous algorithms for multi-task learning with a ``main'' task and an ``auxiliary'' task~\citep{duAdaptingAuxiliaryLosses2020}.
Others have explored variations of this ``gradient surgery''~\citep{yuGradientSurgeryMultiTask2020}.
To our knowledge, such ideas have not yet been suggested or validated for closed-set or open-set SSL.

\textbf{Inspiration from continual learning.}
Our step modification phase is also inspired by the \textit{Transfer-Interference trade-off}~\citep{riemer2018learning, lopez-pazGradientEpisodicMemory2017}.
This trade-off measures whether learning from one example will improve or impair learning on another example.
These works formally define
\emph{transfer} as the case where the inner product of each example's loss gradient with respect to weights is positive, and
\emph{interference} as the case where the inner product is negative.
Other continual learning work also pursues this direction~\citep{chaudhry2018efficient, he2018overcoming, zeng2019continual, farajtabar2020orthogonal}.
We extend this transfer-interference idea to SSL.

% MCH: cut for space. I think verbal descr above is enough.
%Formally, when training with gradient descent, given current network weight $w$, loss function $L$ and two arbitrary distinct samples ($x_{i}$, $y_{i}$), ($x_{j}$, $y_{j}$). 
%\textit{Transfer} occurs when:
%\begin{align}
% \frac{\partial L(x_{i}, y_{i})}{\partial w} \cdot \frac{\partial L(x_{j}, y_{j})}{\partial w} > 0
% \label{def:transfer}
%\end{align}

%While \textit{Interference} occurs when:
%\begin{align}
% \frac{\partial L(x_{i}, y_{i})}{\partial w} \cdot \frac{\partial L(x_{j}, y_{j})}{\partial w} < 0
% \label{def:interference}
%\end{align}

% MCH: cut for space
% MCH: should we move to appendix?
%\textbf{Definition 2: Learning of unlabeled sample harms the learning of labeled sample.} Given current network weight $w$, labeled loss function $L^{L}$, unlabeled loss function $L^{U}$, a labeled sample ($x^{L}_{i}$, $y^{L}_{i}$) and an unlabeled sample $x^{U}_{j}$. 
%\begin{align}
%\frac{\partial L^{L}(x^{L}_{i}, y^{L}_{i})}{\partial w} \cdot \frac{\partial L^{U}(x^{U}_{j})}{\partial w} < 0
% \label{def:ssl_interference}
%\end{align}

% \textbf{Comparison to OOD filtering.}
% Using the Fix-A-Step update, we know that each (stochastic) gradient step could beneficially reduce the labeled set loss, given a small-enough step size $\epsilon$.
% We argue our step direction modification is \emph{safer} than simply learning to downweight individual terms in the unlabeled loss.
% Without care, the latter could still take steps in a problematic non-descent direction of the labeled loss.

\textbf{Simplicity compared to related work.}
We emphasize a key advantage of Fix-A-Step is \emph{extreme simplicity}.
Beyond the modest cost of MixMatch-like augmentation, we compute exactly the same losses and gradients as any standard deep SSL solving Eq.~\eqref{eq:loss-for-consistency-ssl}.
Each possible weight update is straightforward.
Determining which update to use depends only on an inner product, adding negligible runtime cost.
Table 1 suggests Fix-A-Step is favorable to other open-set SSL approaches in its simplicity.
There is no added complexity from extra backward passes, no extra neural networks that must be trained for OOD discrimination, no need for curriculum learning, and no expensive bi-level optimization problem to solve.
Simplicity leads to faster training (App.~\ref{tab:cifar10_runtime_acc}, ~\ref{tab:TMED2_runtime_acc}) and (hopefully) easier adoption.

\textbf{Synergy between Phase 1 and 2.}
One might wonder if our step direction modification in Phase 2 leads to overfitting the labeled loss.
We argue that Phase 1's augmentation should protect against overfitting, and our experiments thus far suggest overfitting is not a major concern.

%concern since regularization techniques such as data augmentation and weight decay are used. Further, the ultimate goal is generalization performance, and whichever lead us to this goal (either from labeled loss or unlabeled loss) is fine.}

\section{EXPERIMENTS ON CIFAR}
\label{sec:results-cifar}
%%% FIG: CIFAR10
\begin{figure*}[!t]
\centering
\begin{tabular}{c c}
\begin{minipage}{.15cm}
\rotatebox[origin=c]{90}{{400 examples/class}}
\end{minipage}
&
\begin{minipage}{.95\textwidth}
\includegraphics[width=\textwidth]{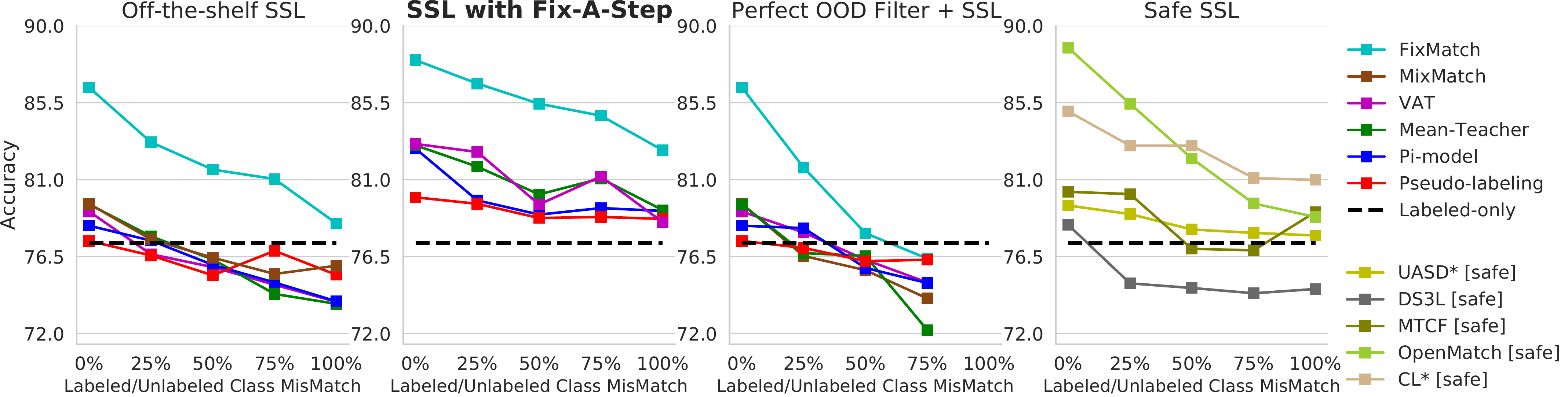}
% \\
% \includegraphics[width=\textwidth]{figures/CIFAR10_400perclass_4columns.pdf}
\end{minipage}
\\[2.3cm] % HACK add bit of whitespace between rows
\begin{minipage}{.15cm}
\rotatebox[origin=c]{90}{{50 examples/class}}
\end{minipage}
&
\begin{minipage}{.95\textwidth}
\includegraphics[width=\textwidth]{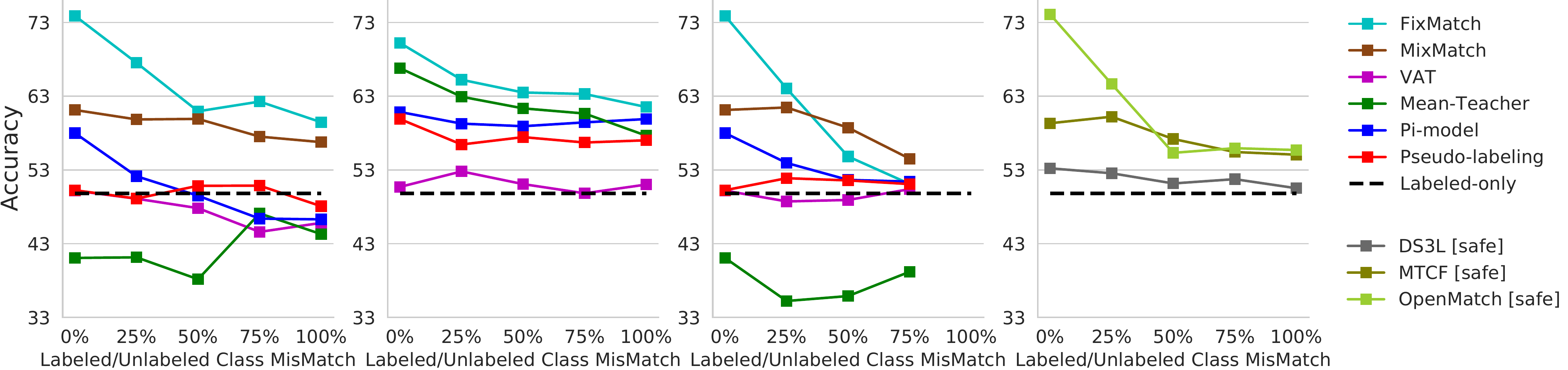}
\end{minipage}
\end{tabular}
\caption{
\textbf{Accuracy on CIFAR-10 6 animal task.}
Accuracy on test images of animals (y-axis) as unlabeled set mismatch (percentage of non-animal classes represented, x-axis) increases.
\emph{Column 1 (from left):} Previous SSL methods trained in standard fashion.
\emph{Col. 2:} SSL methods trained with our Fix-A-Step.
\emph{Col. 3:} SSL methods with perfect OOD filtering of the unlabeled set (removing all non-animal images before training).
\emph{Col. 4:} Previous methods designed for open-set or safe SSL.
UASD and CL (marked $*$) taken from its publication, others from our experiments.
\emph{Top row:} 400 examples/class; \emph{Bottom:} 50 examples/class.
%% MCH cut the below for space
%We plot results of our own experiments with carefully-matched implementations when possible; \todo{lines marked with asterisk $*$ indicate where we copied accuracy numbers from previous publications.}\todo{We directly apply the same hyper-parameters from 400 examples/class experiments to 50 examples/class experiments} 
%Following the experimental design of \citet{oliverRealisticEvaluationDeep2018}, labeled set contains 6 animal classes.
%Fraction of unlabeled set representing non-labeled classes increases from 0-100\%.
}%endcaption
\label{fig:results-cifar10}
\end{figure*}

Our open source code uses PyTorch \citep{paszke2019pytorch} and allows reproducing each experiment (see App.~\ref{app:experiment_details}).
Following \citet{oliverRealisticEvaluationDeep2018}, for all methods we use the same Wide ResNet-28-2  \citep{zagoruyko2016wide}, apply standard augmentation (random crops, flips) on the labeled set, and regularize via weight decay.

\textbf{Hyperparameters.}
All baselines use well-tuned hyperparameters for CIFAR-10 suggested by previous work ~(see App.~\ref{app:hyperparameter_list}).
If a baseline underperformed, we retuned to maximize validation set accuracy.
To be sure our reported gains are meaningful, we did \emph{no hyperparameter tuning at all} for Fix-A-Step, fixing $\alpha=0.5, \tau=0.5$ throughout and inheriting other hyperparameters from the base SSL method.

\textbf{SSL baselines.} 
We compared to 6 closed-set SSL methods (Pi-Model, Mean-Teacher, Pseudo-label, VAT, MixMatch, and FixMatch) as well as the baseline that minimizes labeled loss $\ell^L$ on the labeled set (``labeled-only'').
We also compare to 5 state-of-the-art methods intended for open-set/safe SSL: UASD~\citep{chenSemiSupervisedLearningClass2020}, DS3L~\citep{guoSafeDeepSemiSupervised2020}, MTCF~\citep{yuMultiTaskCurriculumFramework2020}, OpenMatch~\citep{saitoOpenMatchOpensetConsistency2021} and Curriculum-labeling~\citep{cascante-bonillaCurriculumLabelingRevisiting2021}.
If possible, we use our own implementations of baselines, ensuring architectures, training, and hyperparameters are comparable and reproducible.
If a result is copied from another paper, we mark with an asterisk ($*$).

\textbf{Training.}
Following choices in original implementations, each method is trained using either Adam with fixed learning rate or SGD with a \emph{cosine-annealing schedule} for learning rate~\citep{sohnFixMatchSimplifyingSemisupervised2020}.
%We used Adam with fixed learning rate for MixMatch following implementation from \cite{berthelotMixMatchHolisticApproach2019}.
%We used SGD with a \emph{cosine-annealing schedule} for learning rate for Pi-Model, Mean-Teacher, Pseudo-label and FixMatch following implementation from \cite{sohnFixMatchSimplifyingSemisupervised2020}.
We found cosine-annealing and a slow linear ramp-up schedule for the unlabeled-loss-weight $\lambda$ particularly helpful for several baselines (see App.~\ref{app:experiment_details}).
Each training run used one NVIDIA A100 GPU.

\subsection{CIFAR-10 Protocol and Results}

\textbf{6-animal task for CIFAR-10.}
We pursue the ``6-animal'' task designed by \citet{oliverRealisticEvaluationDeep2018} to artificially create unlabeled sets at different levels of mismatch with the labeled set.
We build a labeled set of the 6 animal classes (dog, cat, horse, frog, deer, bird) in CIFAR-10, across two training set sizes: 50 labeled images per class and 400 per class.
We form an unlabeled set of ${\sim}4100$ images/class from 4 selected classes, some animal and some non-animal (car, truck, ship, airplane).
The percentage of non-animal classes is denoted by $\zeta$.
If $\zeta=0\%$, we recover the standard ``closed-set'' SSL setting.
At $\zeta=100\%$, the unlabeled set has no classes in common, and the OOD-filtering paradigm suggests that we should ignore the unlabeled set entirely.
For details on the unlabeled set construction, see App. \ref{tab:CIFAR10_Class_MisMatch_Description}.
%see App.~\ref{app:CIFAR10_Class_Mismatch_description}.

% \textbf{Hyperparameters}.
% All baselines use hyperparameters suggested by previous work for the CIFAR-10 6 animal task~(see App.~\ref{app:hyperparameter_list}).
% In rare cases, if a baseline underperformed we retuned values to maximize validation set accuracy.
% We did \emph{no hyperparameter tuning} for Fix-A-Step, fixing $\alpha=0.5, \tau=0.5$ throughout and inheriting other hyperparameters from the base SSL method.
%This lack of tuning 

\textbf{Results on 6-animal.}
In Fig.~\ref{fig:results-cifar10}, we compare the accuracy of different methods at recognizing the 6 animal classes in the test set, as the mismatch percentage $\zeta$ increases.
Across two different training set sizes (rows), we compare 4 different training scenarios (columns, best read left to right): methods trained in the standard way (``off-the-shelf''), methods trained using Fix-A-Step, methods trained in the standard fashion but with \emph{perfect OOD filtering} applied to the unlabeled set before training so that only known-class samples remain, and methods intended for safe SSL.
The \emph{perfect OOD filtering} column essentially shows the best-possible case for methods under the OOD-is-harmful paradigm.

We highlight several findings from Fig.~\ref{fig:results-cifar10}:
% \textbf{1. Existing ``safe'' SSL methods perform little better than labeled-set-only.}
% In the right-most column, ``safe'' SSL methods (UASD, DS3L, MTCF) roughly match or fall below the dashed line above $\zeta=25\%$.
% FixMatch, which was not intended to be ``safe'', dominates all 3 methods.
% % \todo{(DISCUSS THIS SENTENCE?) This suggests that augmentation may be critical to success.}

\textbf{1. Fix-A-Step improves all SSL methods in almost all settings.}
Despite its relative simplicity, Fix-A-Step is quite effective, as seen in the raised accuracies from the first to the second column across almost all methods and $\zeta$ values.
Fix-A-Step with FixMatch base outperforms all other safe SSL methods (4th col.) for all mismatch levels $\zeta > 0\%$.
In Fig.~\ref{fig:cifar10_uncertainty}, we further demonstrate that Fix-A-Step's gains are \emph{robust} across multiple random train/test splits.

\textbf{2. Perfect OOD filtering is not enough.}
The third column shows that perfect OOD filtering delivers underwhelming accuracy compared to Fix-A-Step for all $\zeta>0$. 
% \todo{It can be seen as an upper bound telling us what happens if we only try to filter OOD samples out}.
% The gains of our Fix-A-Step over this best-case filtering suggest that our \emph{OOD-is-helpful} paradigm should be prioritized over filtering.
Our method's gains over perfect filtering suggest that \textit{trying to benefit from OOD samples is more useful than filtering them.}
We suggest several explanations for the poor performance of perfect filtering, such as 
class imbalance even among known classes in the unlabeled set~\citep{kim2020distribution,lai2022smoothed}, %zhao2022dc guo2022class
sensitivity to hyperparameters~\citep{su2021realistic,sohnFixMatchSimplifyingSemisupervised2020}, % chen2020negative
or perhaps how unlabeled data may affect the training via batchnorm ~\citep{zhao2020robust}. 
More work is needed to understand this phenomenon.
%class imbalance even among known classes in the unlabeled set~\citep{zhao2022dc, kim2020distribution,guo2022class, lai2022smoothed},
%sensitivity to hyperparameters~\citep{su2021realistic,chen2020negative,sohnFixMatchSimplifyingSemisupervised2020}, or perhaps how unlabeled data affect the training via batchnorm ~\citep{zhao2020robust}. 

\textbf{3. Fix-A-Step is faster than alternatives.}
For example, in the 400 examples/class $\zeta=50\%$ setting, using Fix-A-Step with a Mean-Teacher base delivers similar accuracy to OpenMatch (81.08 vs 79.62) while requiring \emph{less than half the training time} (22 vs. 47 hr., App ~\ref{tab:cifar10_runtime_acc}).

\subsection{CIFAR-100 Protocol and Results}

% We now consider the larger CIFAR-100 dataset, following the experimental design of \todo{CITE} to artificially create an unlabeled set and assess performance at various levels of labeled/unlabeled mismatch.

\textbf{50-class task for CIFAR-100.}
Using the larger CIFAR-100 dataset, we follow the open-set SSL experimental design of \citet{chenSemiSupervisedLearningClass2020} to create a $\zeta = 50\%$ class distribution mismatch scenario by using classes 1-50 as labeled classes, and classes 25-75 as unlabeled classes. 
To assess a more extreme level of unlabeled set ``contamination'', we further create a 100\% class distribution mismatch scenario: classes 1-50 are labeled classes; classes 51-100 unlabeled. 
%\todo{For both scenario, we form a labeled set containing 100 images per class and an unlabeled set contains 350 images per class. The validation set contains 50 images per class from the labeled classes and the test set contains 100 images per class from the labeled classes.}

%\textbf{Hyperparameters}. All methods use the same hyperparameters as in CIFAR-10 experiments, without any retuning.

%%% FIG: CIFAR100
\begin{figure*}[!t]
\centering
\begin{tabular}{c c}
\includegraphics[width=.46\textwidth]{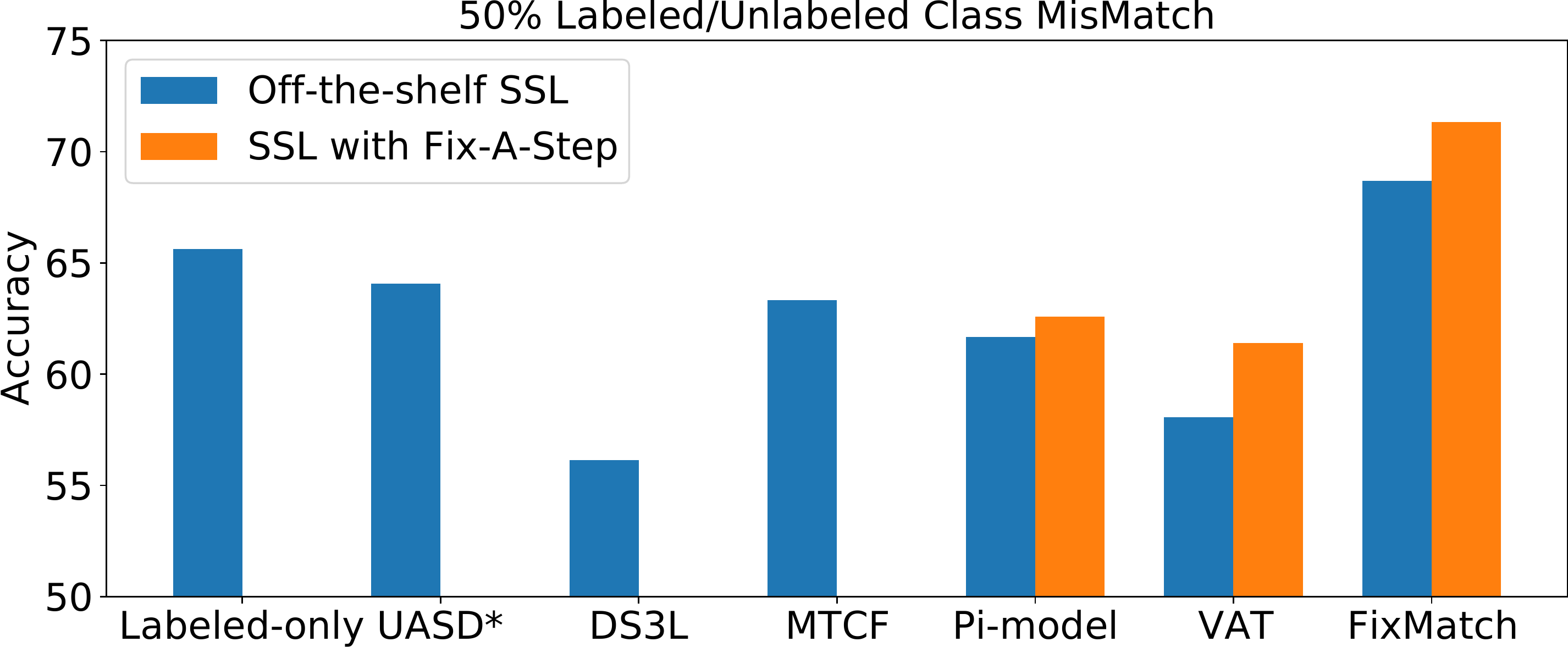}
&
\includegraphics[width=.46\textwidth]{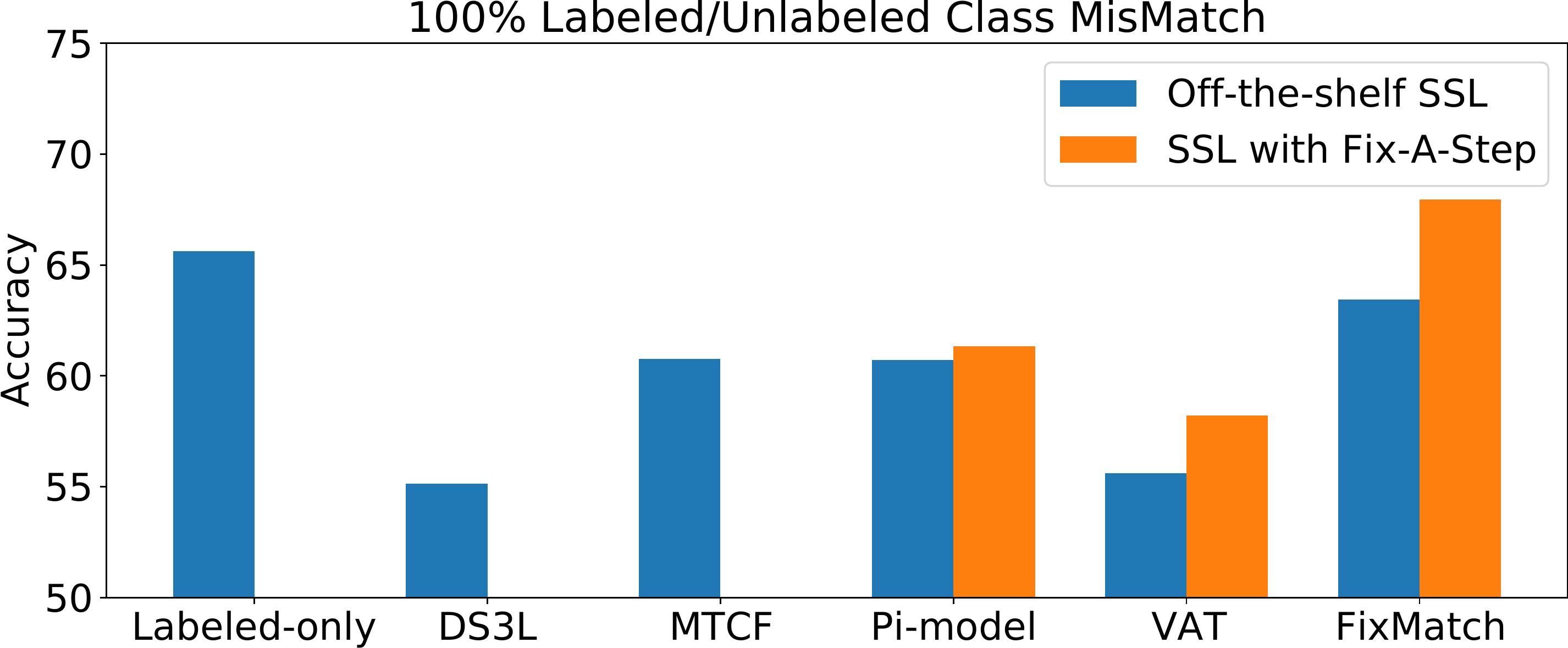}
\end{tabular}
\caption{
\textbf{Accuracy on CIFAR-100 50-class task.}
Each bar represents the accuracy of a method with either off-the-shelf training (blue) or Fix-A-Step (orange).
We try 2 scenarios: 50\% labeled/unlabeled class mismatch (\emph{left panel}) and 100\% class mismatch (\emph{right}).
All numbers were produced by our implementation except those marked $*$ (UASD).
}%endcaption	
\label{fig:results-cifar100}
\end{figure*}

%Every method uses the same hyper-parameters as their corresponding CIFAR-10 experiments.
%\todo{Asterisk $*$ indicate where we copied accuracy numbers from previous publications.}

% \begin{figure*}[!t]
% \centering
% \begin{tabular}{c c c}
% \begin{minipage}{.5cm}
% \rotatebox[origin=c]{90}{{\Large 50 examples/class}}
% \end{minipage}
% &
% \begin{minipage}{.3\textwidth}
% \includegraphics[width=\textwidth]{figures/FixAStep_CIFAR10.pdf}
% \end{minipage}
% \end{tabular}
% \caption{
% SSL accuracy on larger CIFAR-100 (y-axis) as unlabeled set mismatch increases (x-axis), using XYZ instances/class.
% \emph{Left:} off-the-shelf SSL and safe SSL methods.
% \emph{Center:} SSL methods trained with Fix-A-Step.
% Following the experimental design of \citet{oliverRealisticEvaluationDeep2018}, labeled set contains 6 animal classes.
% Fraction of unlabeled set representing non-labeled classes increases from 0-100\%.
% }%endcaption
% \label{fig:results-cifar10}
% \end{figure*}

\textbf{Results on CIFAR-100 50-class.}
Fig.~\ref{fig:results-cifar100} compares ``off-the-shelf'' SSL methods using standard training (blue bars) and Fix-A-Step (orange).
We see consistent gains at both 50\% and 100\% mismatch,
even without tuning hyperparameters.
% Other methods (UASD, DS3L, MTCF) underperform, perhaps due to untuned hyperparameters.
%\todo{No other ``safe'' SSL method (UASD, DS3L, MTCF) can beat even the labeled-set-only baseline.}
%\emph{CIFAR-100}: For all methods used in CIFAR-100 experiments, we directly used the hyperparameters from their corresponding CIFAR-10 experiments, and do not re-tune any hyperparameters.

\subsection{Ablations and Sensitivity Analysis}

\textbf{Ablations.}
We quantify how each of Fix-A-Step's two key components (Augmentation and Gradient step modification) perform in isolation.
Tab.~\ref{tab:results_cifar10_ablation} compares accuracy on the 6 animal task at 400 examples/class and $\zeta = 100\%$. 
Gradient step modification alone increases accuracy around 0.5 to 1.5\% across five base SSL methods.
Augmentation alone increases accuracy around 2.5 to 4.5\%. \emph{When combined, we consistently see the largest gains}. Although we didn't tune hyperparameters, we expect enlarging batch size may lead to more gain from gradient step modification, since it gives less noisy estimates of the gradient alignment.
For further results at other mismatch levels, see App.~\ref{app:cifar}. 
%Note that even though the augmentation seems more effective than the gradient step modification, adding gradient step modification on top of augmentation (A\&G) consistently give better result than augmentation alone.

\textbf{Sensitivity analysis.} 
There are two hyperparameters unique to Fix-A-Step:
sharpening temperature $\tau{>}0$ and the Beta shape $\alpha {>}0$.
For simplicity, we set $\alpha{=}0.5$ and $\tau{=}0.5$ throughout. % our experiments.
Since deep SSL is often sensitive to hyperparameters, we further analyse other possible choices: $\alpha {\in} \{0.5, 0.75\}$ and $\tau {\in} \{0.5, 0.95\}$.
% Since deep SSL is often sensitive to hyperparameters~\citep{su2021realistic}, we further analyse other possible choices: $\alpha {\in} \{0.5, 0.75\}$ and $\tau {\in} \{0.5, 0.95\}$.
%We choose 2 common $\alpha$ values 0.5 and 0.75, 2 common $\tau$ values 0.5 and 0.95, resulting in total of 4 $\alpha$ and $\tau$ combinations.
%We run Mean-Teacher with Fix-A-Step for the 4 combinations of $\alpha$ and $\tau$ across 5 random split of the data. 
Fig.~\ref{fig:cifar10_sensitivity} shows that Fix-A-Step delivers consistent and similar accuracy gains across \emph{all} tested $\alpha,\tau$ settings, and thus does not appear overly sensitive.

%%% TABLE OF ABLATIONS
\setlength{\tabcolsep}{.08cm}
\begin{table}[!t]
\centering
\begin{tabular}{l | r  r  r  r r}
& Pi-Model & MT & VAT & Pseudo & FixMatch
\\ \hline
off-the-shelf &
	73.90 & 73.75 & 73.87 & 75.45 & 78.45
\\
+G only &
	74.50 & 74.33 & 75.35 & 75.92 & 79.73
\\
+A only &
	77.25 & \textbf{78.38} & \textbf{77.87} & \textbf{77.88} & 81.53
\\
+A\&G (ours) &
	\textbf{79.18} & \textbf{79.23} & \textbf{78.52} & \textbf{78.72} & \textbf{82.73}

\end{tabular}
    \caption{
    \textbf{Ablations for CIFAR-10 6 animal task}, reporting accuracy for each SSL method (columns) if we only use our augmentation (+A), only use our gradient step modification (+G), or use the combination (+A\&G) that defines Fix-A-Step.
    We \textbf{bold} the best result and all others within 1 percentage point.
    Setting: 400 examples/class, $\zeta=100\%$.
%\vspace{-0.5cm}
}%endcaption
    \label{tab:results_cifar10_ablation}
\end{table}

\section{EXPERIMENTS ON HEART2HEART}
\label{sec:results-tmed}
%\subsection{Uncurated SSL for Echocardiogram Views}

%%% FIG: TMED2
\begin{figure*}[!t]
\centering
\begin{tabular}{c c c}
TMED2 (Boston, USA; 4 views) & Transfer to Unity (UK; 3 views) & Transfer to CAMUS (France, 2 views)
\\
\includegraphics[width=.32\textwidth]{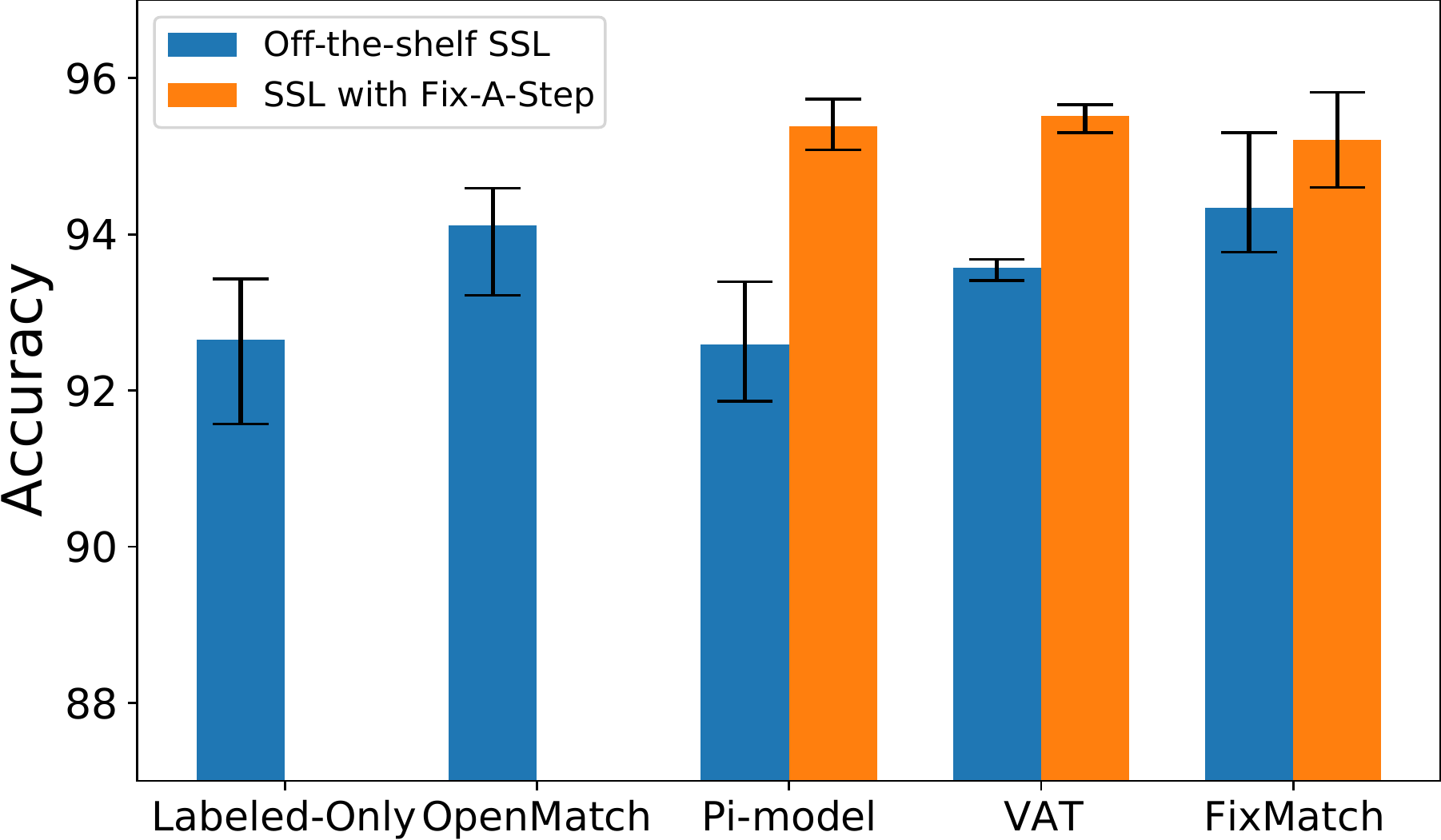}
&
\includegraphics[width=.32\textwidth]{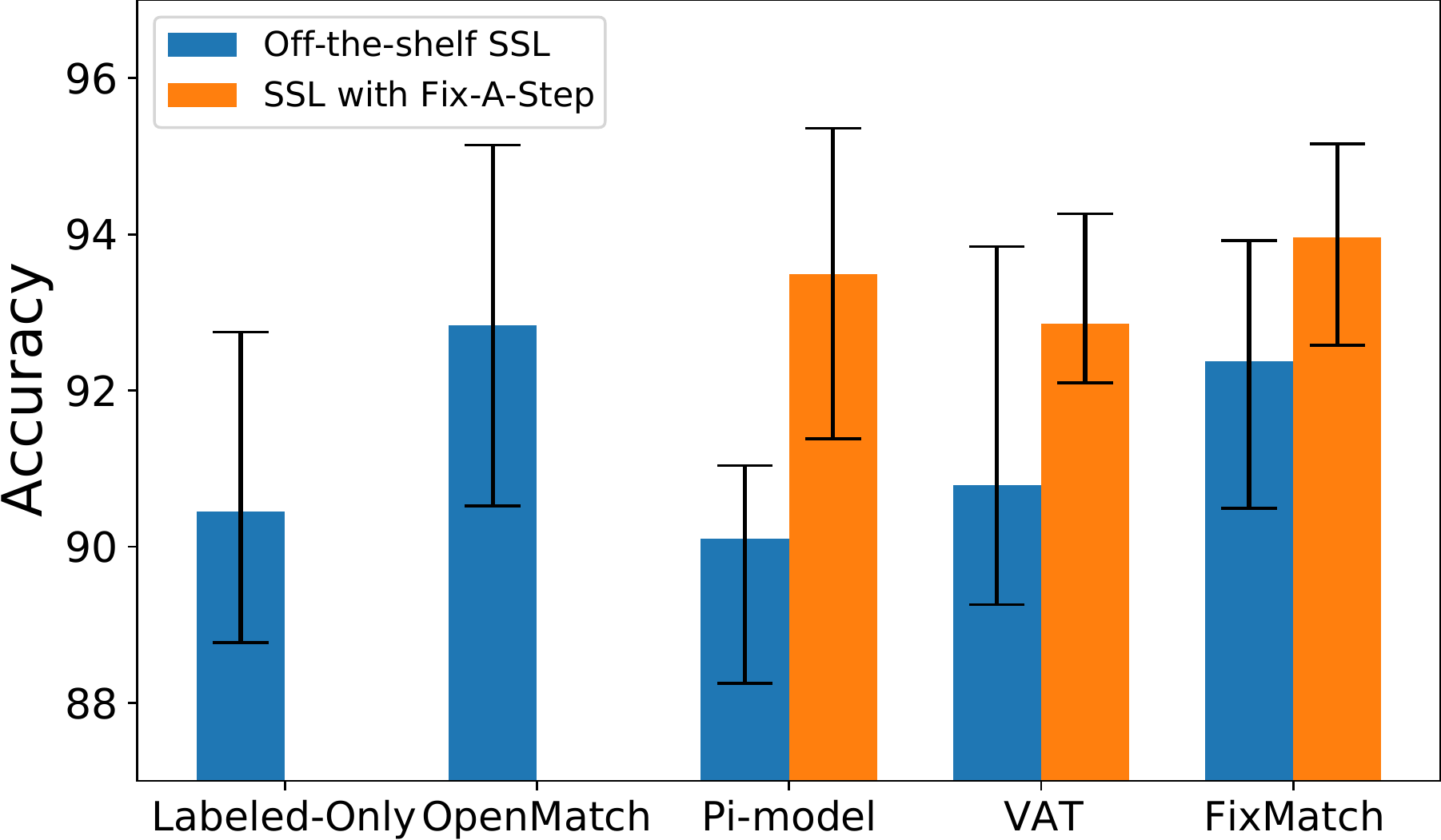}
&
\includegraphics[width=.32\textwidth]{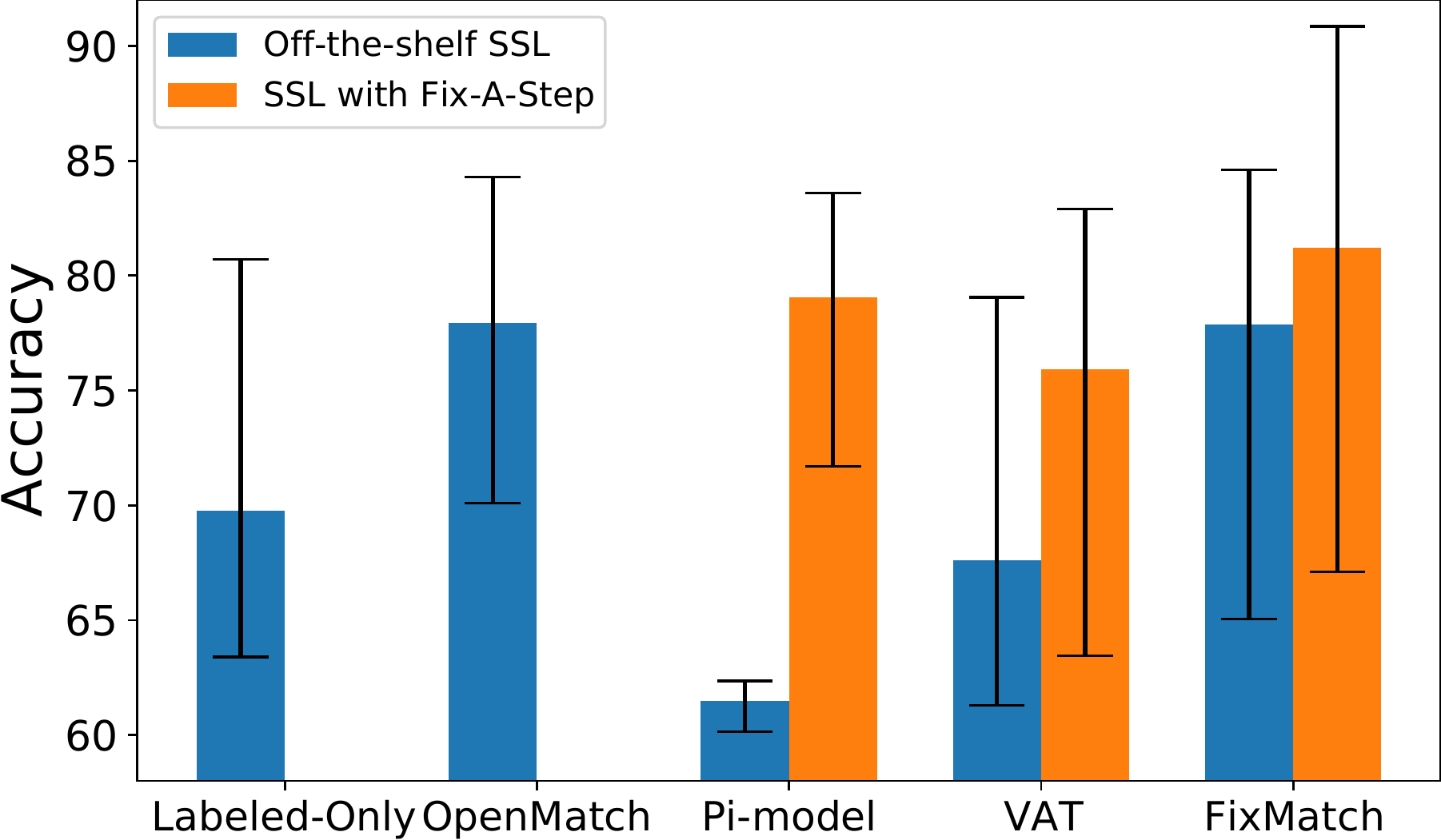}
\end{tabular}
\caption{
\textbf{Balanced accuracy for echocardiogram view classification (Heart2Heart benchmark).}
Methods are trained on TMED-2 images to distinguish 4 view types: PLAX, PSAX, A2C, and A4C.
TMED2's 353,500 image unlabeled set is \emph{uncurated}, representing a superset of possible view types including the 4 known classes.
Bar height gives mean balanced accuracy across 3 models trained on different splits of TMED-2 (error bars indicate min/max).
\emph{Left:} Evaluation on heldout TMED-2 images.
\emph{Center:} Evaluation of TMED-2-trained classifiers on PLAX, A2C, and A4C images from Unity dataset (17 sites in the UK). %7513
\emph{Right:} Evaluation of TMED-2-trained classifiers on A2C and A4C views from CAMUS dataset (1 site in France).
% representing A2C and A4C views from patients at a French hospital.
}%endcaption	
\label{fig:results-heart2heart-view}
\end{figure*}

In pursuit of realistic evaluation, we consider a reproducible, clinically-relevant SSL task that we call \emph{Heart2Heart}.
% using open access medical data.
The key question is this: can we transfer classifiers trained on ultrasound images of the heart from one hospital system to new heart images from unrelated hospitals in other countries.
For training, we use the \emph{Tufts Medical Echocardiogram Dataset 2} (TMED-2)~\citep{huangTMEDDatasetSemiSupervised2022,huangNewSemisupervisedLearning2021}, collected in Boston, USA.
TMED-2 has a small labeled set of echocardiogram studies and a larger \emph{uncurated} unlabeled set.
% of 5486 studies (353,500 images).
Thanks to common device standards, these images are interoperable with two other datasets of ``echo'' images: \emph{Unity} from 17 hospitals in the UK~\citep{howardAutomatedLeftVentricular2021} and \emph{CAMUS} from a hospital in France~\citep{leclercDeepLearningSegmentation2019}.
We emphasize that all datasets are deidentified and accessible to any academic researcher.
% for non-commercial purposes.

% MCH: agree we can cut this
%\todo{IS THIS SENTENCE REDUNDANT SINCE WE ALREADY SAID ABOUT RELEASING CODE. We will release code so this \emph{Heart2Heart} evaluation of SSL can be readily reproduced for future SSL methods.}

\textbf{Classification task: View type of 2D TTE image.}
Trans-thoracic echocardiography (TTE) is a gold-standard way to non-invasively capture the heart's 3-dimensional anatomy for measurement and diagnosis.
A human sonographer wields a handheld transducer over the patient’s chest at different angles in order to provide clear views of each facet of the heart.
%We focus on \emph{2-dimensional} (2D) view types, leaving other modalities such as Doppler profiles or m-mode imaging for future work.
A routine TTE scan of a patient, called a \emph{study}, produces many images (median=68, 10-90th percentile range=27-97 in TMED-2), each showing a canonical 2D view of the heart.
No view type annotation is recorded with any image.
In later analysis, clinicians manually search over all images to find a desired view type.
Automated interpretation of echocardiograms must also be able to pick out specific view types before any useful measurements or diagnosis can be made, making \emph{view classification} a prediction task with potential clinical impact~\citep{madaniFastAccurateView2018,huangNewSemisupervisedLearning2021}.

%This tedio difficulty motivates the need for a \emph{view classifier}. Can we reliably find the views relevant to AS from the dozens of images in a typical study?

%\emph{multiple} short videos of the heart, each depicting a potentially different anatomical view throughout the cardiac cycle.
%From each short video, we can extract a representative image.

TMED-2 provides set of labeled images of four specific view types, known as PLAX, PSAX, A2C, and A4C, gathered from certified annotators.
% at great expense.
Reliably identifying these views would be particularly useful for key valve disease diagnostic tasks~\citep{huangNewSemisupervisedLearning2021, wessler2023automated}.
TMED-2 also contains a truly \emph{uncurated} unlabeled set of 353,500 images from routine TTEs from 5486 patient-studies.
At least 9 canonical view types frequently appear in routine TTEs~\citep{mitchellGuidelinesPerformingComprehensive2019}, so this unlabeled set should contain extra classes not in the labeled set. However, this view classification problem on TMED-2 is more than just ``open-set'' SSL, because TMED-2 labeled and unlabeled sets were not identically sampled from the same patient population, which leads to some modest feature differences. Unlabeled echos come from all available files for convenience, while the the labeled set deliberately oversamples patients with a valve disease called aortic stenosis (AS).
About 50\% of all patients in the labeled set have severe AS, compared to less than 10\% in the general population.
For severe AS patients, PLAX and PSAX images will show heavier calcification (thickening) of the aortic valve.

\textbf{Protocol.} 
Averaging over TMED-2's recommended 3 splits, 
we train each SSL method on images from 56 labeled studies as well as all unlabeled studies (353,500 images). 
We report \emph{balanced accuracy} on each split's test set of 120 studies ($\sim$2104 images).
% per split).
% count of the 3 different splits in TMED2 differ by less than 10\%, see \cite{huangTMEDDatasetSemiSupervised2022} for more details}.
%To further assess generalization to new hospitals, we apply the classifier to two other public datasets that contain view-labels for echocardiograms, both gathered on a different continent than TMED.
We then assess \emph{generalization} of these Boston-based classifiers to images from European hospitals.
We report balanced accuracy on 7231 available PLAX, A2C, and A4C images from Unity,
as well as all 2000 images (A2C and A4C views) in CAMUS.

%\subsection{Results: Heart2Heart View Classification}
\label{sec:results-tmed}
\textbf{Results on Heart2Heart.}
Fig.~\ref{fig:results-heart2heart-view} shows classifier performance on held-out data from all 3 datasets.
TMED-2 evaluations (first panel) show that our Fix-A-Step procedure yields gains across all tested SSL methods (Pi-Model, VAT, FixMatch). Fix-A-Step helps all three methods convincingly outperform the labeled-only baseline.
Compared to OpenMatch, a state-of-the-art safe SSL method, Fix-A-Step yields better accuracy while being much simpler. 
%Similar to our finding from CIFAR-10 experiment, w
We also find that Fix-A-Step delivers competitive accuracy considerably faster ($\sim$2-3x speedup, See App. ~\ref{tab:TMED2_runtime_acc}).

% only do the SSL methods with Fix-a-Step perform significantly better than the fully supervised baseline, but all three SSL methods with Fix-a-Step outperform the original method without the addition of Fix-a-Step. 

External evaluation on Unity and CAMUS (Fig.~\ref{fig:results-heart2heart-view} panels 2-3) show that these gains transfer to new hospitals.
Each tested SSL method performs better with Fix-A-Step than standard training.
Across splits we see larger performance variation on Unity and CAMUS than on TMED-2, which highlights the difficulty of generalizing across hospitals as well as importance of external validation. 
All methods perform worse on CAMUS than other datasets; see App.~\ref{app:heart2heart} for further investigations.
Overall, this Heart2Heart benchmark task shows the promise of Fix-A-Step to deliver gains from unlabeled data that generalize better than alternatives.

% MCH :cut for space
%\todo{Note that our proposed Heart2Heart benchmark is not only suitable for studying class contamination in SSL but aslo the class imbalance problem ~\citep{kim2020distribution,guo2022class, lai2022smoothed}, since the uncurated TMED2 unlabeled set contains not only unknown classes but also classes with imbalanced frequencies (even extreme imbalance). With such benchmark we can more reliable measure whether the class-imbalanced algorithms generalize.}
%Thus, Fix-a-Step can help the model generalize to datasets that our model has not been trained on. 

%For a discussion on reasons why our model underperforms on CAMUS in general and the data transformation experiments we did, look in appendix \todo{TODO}

\section{DISCUSSION}
In summary, this paper makes three contributions to deep SSL image classification.
First, we argue that uncurated or OOD data in the unlabeled set can be \emph{helpful}, and should not merely be filtered out.
Experiments in Fig.~\ref{fig:results-cifar10} show that even with perfect OOD filtering most SSL methods deliver underwhelming accuracy gains.
% Second, building on insights from multi-task learning~\citep{duAdaptingAuxiliaryLosses2020}, our Fix-A-Step SSL training procedure uses gradient step modifications that prioritize the labeled-set loss, leading to effective SSL that is substantially simpler than alternatives (no new loss terms or extra neural networks).
Second, we introduce a new training procedure called Fix-A-Step that achieves state-of-the-art SSL performance on uncurated unlabeled sets while being faster and simpler (no new loss terms or extra neural nets).
%These methods hold substantial promise for real-world imaging datasets where labels are scarce.
Finally, we hope our new Heart2Heart benchmark for SSL evaluation inspires robust studies of clinical model transportability across global populations.

\textbf{Limitations.}
Our work's exclusive focus is image classification.
More work is needed to try Fix-A-Step on other data types like time series.
%Fix-A-Step could apply to other types of data.
% \todo{Our estimate of the cosine similarity between two gradients can be noisy. 2. does not guarantee maximizing transfer but only drop the worst case 3. Simliar to other works that use gradient similarity to measure transfer and interference, we are making the assumption that the unlabeled loss harming the learning of the labeled loss locally (at step t) implies it will affect the learning of the labeled loss globally. In theory however this is not neccessaritly the true.} 
Our experiments on artificial unlabeled sets in Sec.~\ref{sec:results-cifar} focused exclusively on mismatch in the \emph{labels}. We did not systematically explore how shifts in the features $x$ between the labeled and unlabeled set impact performance, though we do emphasize that TMED-2's uncurated unlabeled set likely has such shifts due different acquisition criteria (see Sec.~\ref{sec:results-tmed}).
Fix-A-Step's phase 2 step modification does not guarantee accuracy gains, only protects against possible deterioration.
Omitting unlabeled loss gradients because of harm to a minibatch labeled loss may miss a chance to improve accuracy globally. Recently, \citet{schmutz2022don} found that the optimal choice of the unlabeled loss coefficient $\lambda$ depends on the covariance matrix between $g^L$ and $g^U$, which might provide another way to consider the step modification phase.
% the learning of the labeled loss locally implies it will affect the learning of the labeled loss globally, which might not necessarily be the case.
%Further, similar to prior works that explore such idea ~\citep{yuGradientSurgeryMultiTask2020, duAdaptingAuxiliaryLosses2020}, we are making the assumption that the unlabeled loss harming the learning of the labeled loss locally implies it will affect the learning of the labeled loss globally, which might not necessarily be the case.}
%is also critical; more work is needed.
% whose feature vectors have fixed dimension (required by MixUp).
%More work is needed on alternative data types or multi-modal SSL.

\textbf{Impact statement.}
Work on SSL is often  motivated by its promise  in medical imaging~\citep{huangNewSemisupervisedLearning2021,madaniDeepEchocardiographyDataefficient2018}.
Our Heart2Heart evaluation shows a proof-of-concept for generalization of ultrasound view classifiers across hospitals.
More work is needed to rigorously assess generalizability and translate to improved patient care.
Extra effort is needed to avoid widening current disparities~\citep{celiSourcesBiasArtificial2022}, as the data sources in our Heart2Heart benchmark do not reflect the geographic and racial diversity of many patient populations.
%While all images in Heart2Heart are completely de-identified, we stress the responsibility we carry as researchers to protect the best interests of the individuals who contributed data.
%and never attempt to reidentify patients.

\textbf{Outlook.}
Fix-A-Step is a promising new first-line approach to SSL that can unlock the promise of uncurated unlabeled sets.
We hope future work explores  augmentation and step direction further, 
while extending our focus on \emph{simplicity}, \emph{reproducibility}, and possible benefits of OOD images. 
%\todo{Our gradient step modification rules are simple, and we believe better rules can be developed to further improve the performance of open-set SSL.}

\subsubsection*{Acknowledgements}
We acknowledge financial support from the Pilot Studies Program at the Tufts Clinical and Translational Science Institute (Tufts CTSI NIH CTSA UL1TR002544) and computing infrastructure support from NSF OAC CC* 2018149.
Author B. W. was supported in part by K23AG055667 (NIH-NIA).
Author M.-J. S. was supported by the Tufts Summer Scholars Program.

\renewcommand{\bibsection}{\subsubsection*{References}}
\bibliographystyle{abbrvnat}
\bibliography{refs_from_zotero.bib,refs_manual.bib}

\clearpage
\appendix

%% Config Table-of-Contents to track the sections of the appendix
%\addtocontents{toc}{\protect\setcounter{tocdepth}{0}}

\counterwithin{table}{section}
\setcounter{table}{0}
\counterwithin{figure}{section}
\setcounter{figure}{0}
\counterwithin{algorithm}{section}
\setcounter{algorithm}{0}

%% Use ONE counter for all figs and tables to give unique identifiers in supplement
\makeatletter 
\let\c@table\c@figure
\let\c@lstlisting\c@figure
\let\c@algorithm\c@figure
\makeatother

%\thispagestyle{empty}
 
% For one-column format, uncomment the following:
\onecolumn

\begin{center}
\Large Supplementary Material
\end{center}
~\\
~\\
In this supplement, we provide:
\begin{itemize}
  \setlength\itemsep{0em}
\item Sec.~\ref{app:cifar}: CIFAR Experiments: Further Details, Results, and Analysis 
\item Sec.~\ref{app:heart2heart}: Heart2Heart Experiments: Further Details, Results, and Analysis
\item Sec.~\ref{app:methods}: Methods Supplement: Algorithms for \textsc{Aug+SoftLabel} and \textsc{MixMatchAug}
\item Sec.~\ref{app:related_work}: Related Work Supplement: Further Discussion and Analysis 
\item Sec.~\ref{app:hyperparameter_list}: Reproducibility Supplement: Hyperparameters, Settings, etc.
\end{itemize}

\section{CIFAR EXPERIMENTS: Details, Results, and Analysis}
\label{app:cifar}

\subsection{CIFAR-10 6-animal task mismatch description}

In Table~\ref{tab:CIFAR10_Class_MisMatch_Description} we define which classes form the labeled and unlabeled set at each level of mismatch for the CIFAR-10 6 animal task. This exactly follows ~\citet{oliverRealisticEvaluationDeep2018} and creates \emph{more challenging} scenarios than other ``mismatch'' tasks on CIFAR-10 tried previously (for example, ~\citep{saitoOpenMatchOpensetConsistency2021} examine a case with all 10 classes in the unlabeled set).

\setlength{\tabcolsep}{.8cm}
\begin{table}[!h]
\centering
\begin{tabular}{l | r | r  }
& Labeled set & Unlabeled set
\\ \hline
$\zeta=0\%$ & {Bird, Cat, Deer, Dog, Frog, Horse} & {Deer, Dog, Frog, Horse}
\\
$\zeta=25\%$ & {Bird, Cat, Deer, Dog, Frog, Horse} & {\textbf{Airplane}, Dog, Frog, Horse}
\\
$\zeta=50\%$ & {Bird, Cat, Deer, Dog, Frog, Horse} & {\textbf{Airplane}, \textbf{Car}, Frog, Horse}
\\
$\zeta=75\%$ & {Bird, Cat, Deer, Dog, Frog, Horse} & {\textbf{Airplane}, \textbf{Car}, \textbf{Ship}, Horse}
\\
$\zeta=100\%$ & {Bird, Cat, Deer, Dog, Frog, Horse} & {\textbf{Airplane}, \textbf{Car}, \textbf{Ship}, \textbf{Truck}}
\\

\end{tabular}
    \caption{
    \textbf{Definition of labeled/unlabeled class mismatch scenario in CIFAR-10 6 animal task.} We bolded the non-animal classes in unlabeled set that are not in the labeled set.
    All included classes are represented with equal frequency.
    }%endcaption
    \label{tab:CIFAR10_Class_MisMatch_Description}
\end{table}

\subsection{Ablation Study across different level of contamination}

Expanding on the ablation table in the main paper, in Tab.~\ref{tab:results_cifar10_ablation_all_zeta} we show ablation comparisons (augmentation only, gradient step modification only, or both) across all tested values of the mismatch in labeled-vs-unlabeled class content $\zeta$.

\begin{table}[!h]
	\setlength{\tabcolsep}{.08cm}
    \begin{tabular}{c c}
		{\large Mismatch $\zeta=0\%$}
		&
		{\large Mismatch $\zeta=25\%$}
		\\
		\begin{minipage}{.5\textwidth}
		\resizebox{\textwidth}{!}{%	
			\begin{tabular}{r | r | r | r | r}
    & ~~~~ & +A only & +G only & +A\&G (Fix-A-Step)
\\ 
    Pi-Model & 78.32 & 81.90 & 78.48 & \textbf{82.83} 
\\ 
    Mean-Teacher & 79.57 & \textbf{84.18} & 80.60 & 83.02
\\ 
    VAT & 79.15 & \textbf{83.88} & 79.47 & 83.10
\\
    Pseudo-label & 77.43 & 79.03 & 78.30 & \textbf{79.98}
\\
    FixMatch & 86.40 & 86.35 & 86.17 & \textbf{88.00}
\end{tabular}	
		}
		\end{minipage}
		&
		\begin{minipage}{.5\textwidth}
		\resizebox{\textwidth}{!}{%	
			\begin{tabular}{r | r | r | r | r}
             & ~~~~ & +A only & +G only & +A\&G (Fix-A-Step)
\\ 
    Pi-Model & 77.45 & 79.12 & 78.00 & \textbf{79.80} 
\\ 
    Mean-Teacher & 77.70 & 81.35 & 78.28 & \textbf{87.77}
\\ 
    VAT & 76.65 & 82.27 & 78.35 & \textbf{82.63}
\\
    Pseudo-label & 76.58 & 79.28 & 77.25 & \textbf{79.60}
\\
    FixMatch & 83.20 & 84.44 & 83.85 & \textbf{86.63}
\end{tabular}	
		}
		\end{minipage}
		\\ %%%%%%%%%%%%%%%%%%%%%%%%%%
		\\ %%%%%%%%%%%%%%%%%%%%%%%%%%
		{\large Mismatch $\zeta=50\%$}
		&
		{\large Mismatch $\zeta=75\%$}
		\\
		\begin{minipage}{.5\textwidth}
		\resizebox{\textwidth}{!}{%	
			\begin{tabular}{r | r | r | r | r}
             & ~~~~ & +A only & +G only & +A\&G (Fix-A-Step)
\\ 
    Pi-Model & 76.03 & 78.70 & 76.57 & \textbf{78.97} 
\\ 
    Mean-Teacher & 76.35 & \textbf{82.22} & 78.18 & 80.15
\\ 
    VAT & 75.90 & 79.43 & 77.37 & \textbf{79.56}
\\
    Pseudo-label & 75.42 & 78.33 & 75.65 & \textbf{78.77}
\\
    FixMatch & 81.60 & 83.28 & 81.80 & \textbf{85.45}
\end{tabular}	
		}
		\end{minipage}
		&
		\begin{minipage}{.5\textwidth}
		\resizebox{\textwidth}{!}{%	
			\begin{tabular}{r | r | r | r | r}
             & ~~~~ & +A only & +G only & +A\&G (Fix-A-Step)
\\ 
    Pi-Model & 75.00 & 77.82 & 74.77 & \textbf{79.35} 
\\ 
    Mean-Teacher & 74.33 & 79.63 & 74.52 & \textbf{81.08}
\\ 
    VAT & 74.87 & 80.82 & 75.20 & \textbf{81.20}
\\
    Pseudo-label & 76.85 & 78.38 & 77.15 & \textbf{78.83}
\\
    FixMatch & 81.05 & 83.03 & 81.33 & \textbf{84.75}
\end{tabular}	
		}
		\end{minipage}
	\end{tabular}
	\caption{\textbf{Ablation analysis on CIFAR-10 6 animal task}, examining how accuracy changes for each SSL method if we only use our augmentation (+A), only use our gradient step modification (+G), and use the combination (+A\&G) which constitutes our Fix-A-Step. 
	Each panel shows results for a fixed value of the mismatch percentage $\zeta$ describing the overlap in classes between labeled and unlabeled set.
    For each method, we \textbf{bold} the best result.
    Setting: 400 examples/class.
    }%endcaption
\label{tab:results_cifar10_ablation_all_zeta}
\end{table}

\subsection{Robustness of results across multiple train/test splits}

\begin{figure*}[!h]
\centering
\begin{tabular}{c c}
\includegraphics[width=.35\textwidth]{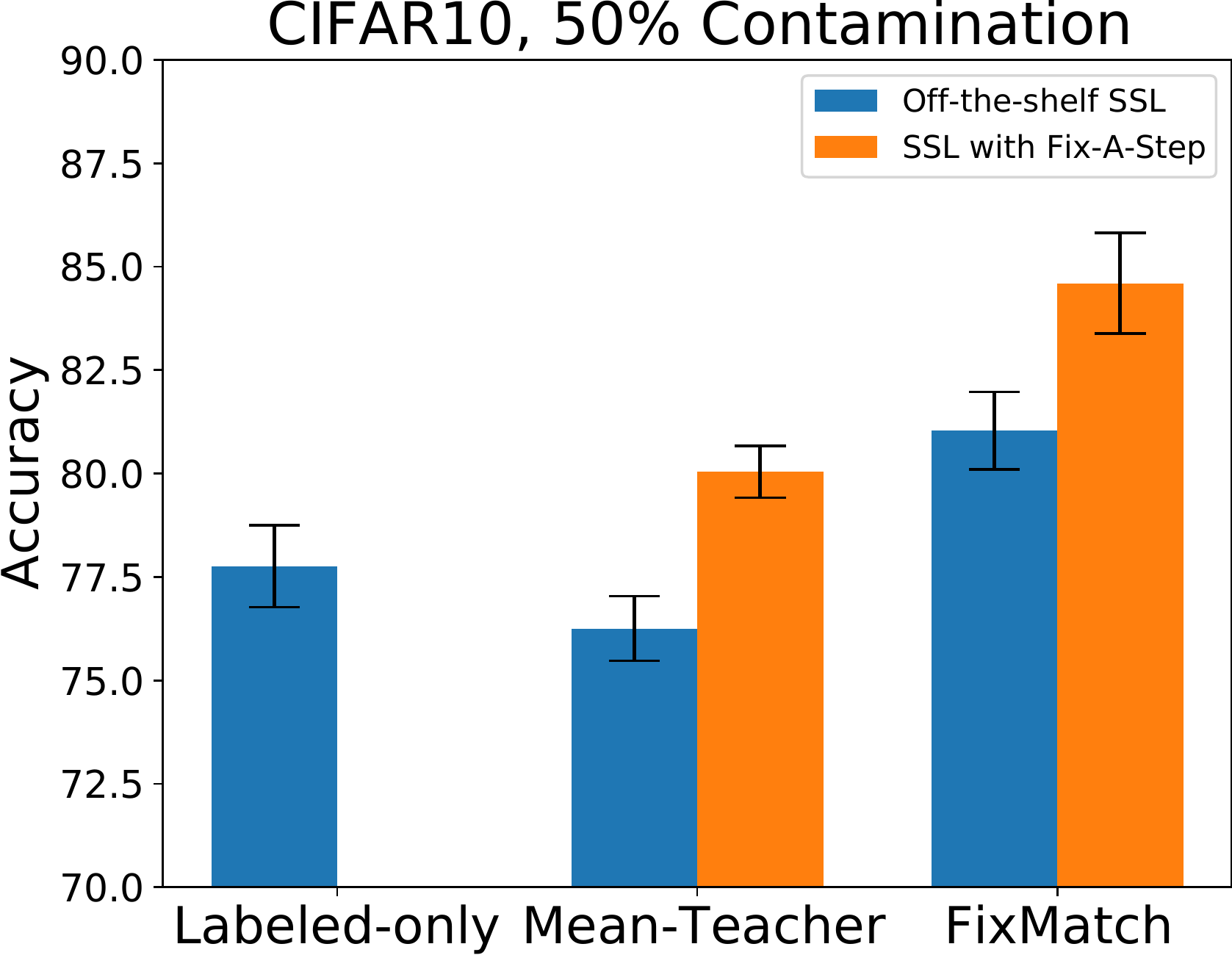}
&
\includegraphics[width=.35\textwidth]{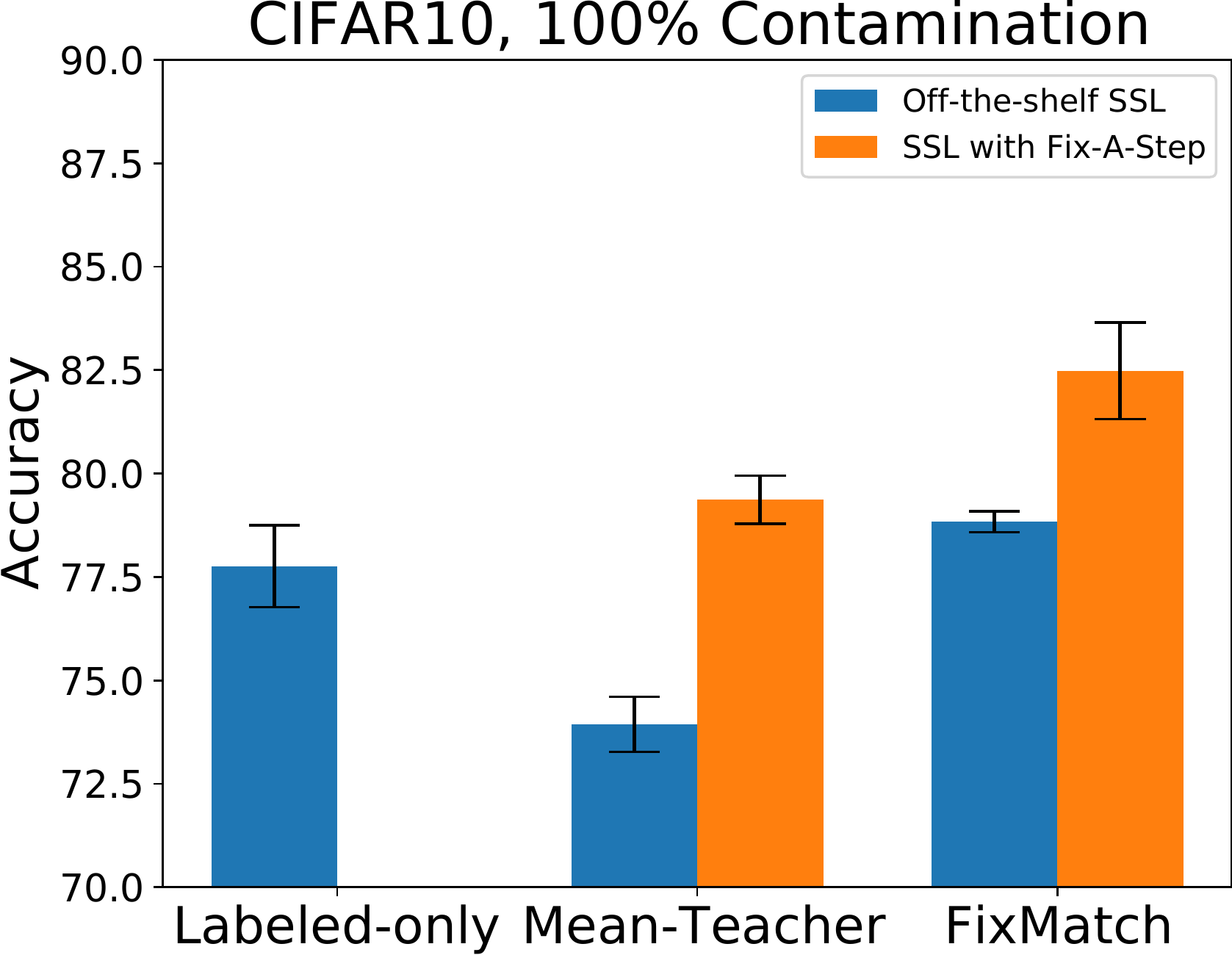}
\end{tabular}
\caption{
\textbf{CIFAR10, 400 examples/class}. Bar height indicates the average across 5 random splits of the data. Error bars show the standard deviation across 5 splits.
}%endcaption	
\label{fig:cifar10_uncertainty}
\end{figure*}

In the main paper, we report results on CIFAR-10 6 animal task across many baselines methods.
For each baseline, we use only one train/test split due to the huge computation required to compare all baselines.
In Fig.~\ref{fig:cifar10_uncertainty}, for a subset of methods we assess the \emph{robustness} of the conclusions of that experiment across multiple separate training/test splits.

We train the labeled-set-only baseline, FixMatch with and without Fix-A-Step, and Mean-Teacher with and without Fix-A-Step for 5 random splits of the data, across two levels of mismatch ($\zeta = 50\%$ and $\zeta = 100\%$).
Results are in Fig.~\ref{fig:cifar10_uncertainty}.
Broadly, we suggest that our conclusion that Fix-A-Step delivers successful accuracy gains holds even across 5 splits: both FixMatch and MeanTeacher show notable gains across both levels of mismatch $\zeta$.
\textbf{Notably, MeanTeacher plus Fix-A-Step appears quite competitive with off-the-shelf FixMatch}, and FixMatch plus Fix-A-Step is the best of all.

\subsection{Sensitivity to hyperparameters}
\begin{figure*}[!h]
\centering
\begin{tabular}{c c}
\includegraphics[width=.42\textwidth]{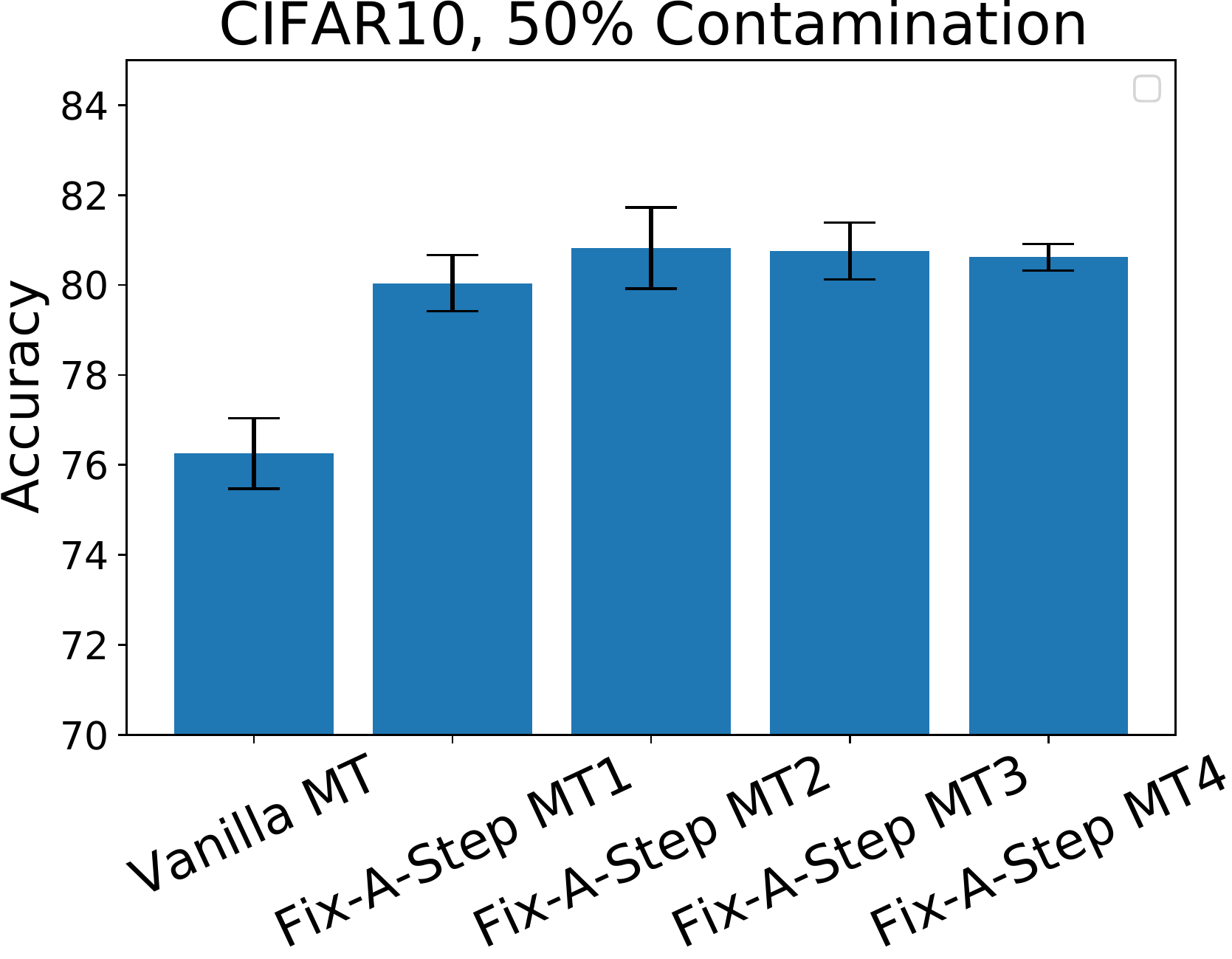}
&
\includegraphics[width=.42\textwidth]{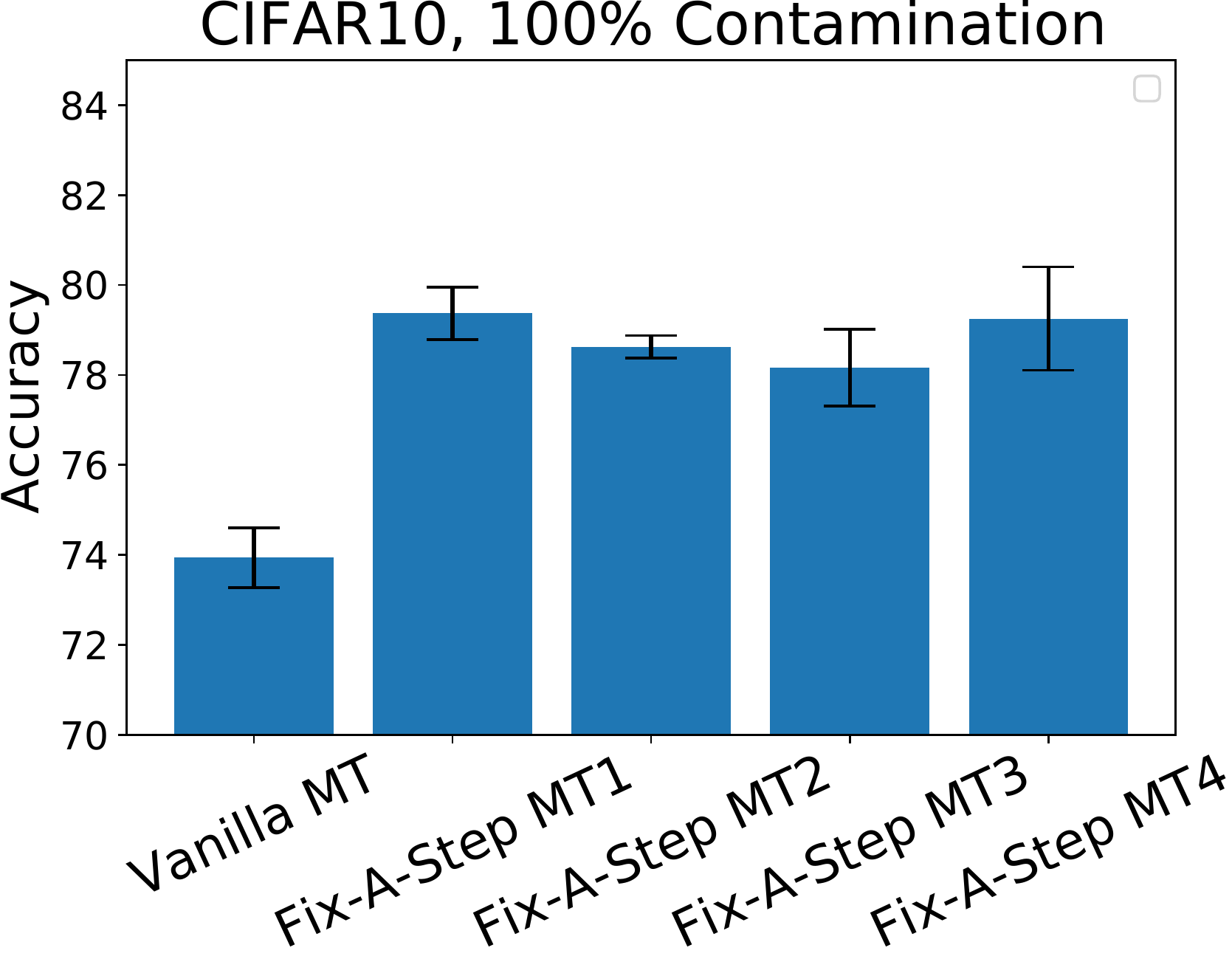}
\end{tabular}
\caption{
\textbf{CIFAR10, 400 examples/class}. \emph{Vanilla MT}: Mean teacher SSL base model. \emph{Fix-A-Step MT1}: MT base Fix-A-Step with $\tau=0.5$ , $\alpha=0.5$, . \emph{Fix-A-Step MT2}: MT base Fix-A-Step with $\tau=0.95$, $\alpha=0.75$. \emph{Fix-A-Step MT3}: MT base Fix-A-Step with $\tau=0.5$, $\alpha=0.75$. \emph{Fix-A-Step MT4}: MT base Fix-A-Step with $\tau=0.95$, $\alpha=0.5$.  Results average across 5 random split of the data. Error bar showing standard deviation across the 5 split.
}%endcaption	
\label{fig:cifar10_sensitivity}
\end{figure*}

Since Deep SSL methods could be sensitive to hyper-parameters, we conduct sensitivity analysis to see how Fix-A-Step behave under different choice of sharpening temperature $\tau$ and Beta distribution shape $\alpha$. We analyzed the performance of Fix-A-Step using a Mean-Teacher base model across several reasonable choices of MixUp parameter $\alpha {\in} \{0.5, 0.75\}$ and sharpening temperature $\tau {\in} \{0.5, 0.95\}$ (totally 4 combinations). (See. Alg.~\ref{alg:FixAStep} for hyperparameter definitions).
Results in Fig.~\ref{fig:cifar10_sensitivity} shows that Fix-A-Step's performance gains over the off-the-shelf (or ``vanilla'') Mean Teacher base do not appear overly sensitive to these hyperparameters.

\vspace{2cm}
\subsection{Comparison of computation cost and performance}
\setlength{\tabcolsep}{.3cm}
\begin{table}[!h]
\centering
\begin{tabular}{c | r r | r r | r r }
\multicolumn{0}{c}{Methods} &\multicolumn{2}{c}{$\zeta=25\%$} &\multicolumn{2}{c}{$\zeta=50\%$} &\multicolumn{2}{c}{$\zeta=75\%$}\\

& Acc & Runtime & Acc & Runtime & Acc & Runtime 
 \\ \hline
MT+Fix-A-Step & 81.77 & 1476 & 80.15 & 1331 & 81.08 & 1333\\
VAT+Fix-A-Step & 82.63 & 1662 & 79.56 & 1691 & 81.20 & 1681\\
OpenMatch & 85.45 & 2728 & 79.56 & 2803 & 79.62 & 3270\\
\end{tabular}
    \caption{
    \textbf{Comparison of runtime and test accuracy on CIFAR-10 6 animal task}. Setting: 400 example/class. Mismatch percentage $\zeta$ describes the overlap in classes between labeled and unlabeled set. Runtime (in minutes) is based on training same number of steps on a Nvidia A100 GPU. 
    }%endcaption
    \label{tab:cifar10_runtime_acc}
\end{table}

\section{HEART2HEART EXPERIMENTS: Details, Results, and Analysis}
\label{app:heart2heart}
\subsection{Comparison of computation cost and performance}
\setlength{\tabcolsep}{.3cm}
\begin{table}[!h]
\centering
\begin{tabular}{c | r r | r  r | r  r }
\multicolumn{0}{c}{Methods} &\multicolumn{2}{c}{split0} &\multicolumn{2}{c}{split1} &\multicolumn{2}{c}{split2}\\

& Acc & Runtime & Acc & Runtime & Acc & Runtime 
 \\ \hline
Pi-model+Fix-A-Step & 95.33 & 233 & 95.08 & 240 & 95.73 & 218\\
VAT+Fix-A-Step & 95.58 & 392& 95.30 & 343 & 95.66 & 356\\
OpenMatch & 94.54 & 1244& 94.59 & 1282 & 93.22 & 879\\

\end{tabular}
    \caption{
    \textbf{Comparison of runtime and test balanced accuracy on TMED-2 view classification task}. Runtime in minutes. Each model is trained on a Nvidia A100. In practice, we found OpenMatch converges slower than alternatives compared, we thus train about 2x more iterations for OpenMatch (otherwise its accuracy performance would be worse).
    }%endcaption
    \label{tab:TMED2_runtime_acc}
\end{table}
% \clearpage

\subsection{Preprocessing TMED-2 data}

We applied for access to the TMED-2 data via the form on the website (\url{https://TMED.cs.tufts.edu}), and downloaded the shared folder of data from the provided cloud-based link after approval.
Images (as 112x112 PNG images) and associated view labels (in CSV files) are readily available in the provided shared folder for download.

\textbf{Train/validation/test splits.}
To form our labeled sets for training, we used the provided train/test splits of the fully-labeled set with the smallest training set size (56 studies available for both training and validation).
While larger labeled training sets are possible, we selected this smaller size as the most compelling use case for SSL.
We wanted to answer the question: how well can we do with very little labeled data but a large pile of unlabeled data.

\textbf{View label selection.}
Among available view labels, we chose PLAX, PSAX, A4C, and A2C as the 4 classes to focus on for our Heart2Heart view type classifier.
The original TMED-2 labeled set, as described in \citet{huangTMEDDatasetSemiSupervised2022}, contains an additional view type label that they called A2CorA4CorOther, which is a super-category that contains possible view types distinct from PLAX and PSAX (including A2C, A4C, and other possible classes like A5C).
For simplicity, we excluded that class in our Heart2Heart experiments.

\subsection{Preprocessing Unity data}

We downloaded the Unity data by going to their website (\url{https://data.unityimaging.net}). Once at their website, go to the 'Latest Data Release' section and download the images.
For the view labels, go to \url{https://data.unityimaging.net/additional.html} and download the csv file under the 'View' section. 

In the Unity dataset, along with PLAX, A2C, and A4C views, there are also A3C and A5C. For the purposes of these experiments, we filtered out all A3C and A5C images.

Disclaimer: These view labels were done by one human so there may be some errors in the labeling. 

The raw Unity data came in .png format, so first we converted all the pngs to a tiff format.
Then we converted them to gray-scale, padded the shorter axis to achieve a square aspect ratio, and resized it to 112 x 112 pixels. 

%Finally, all images and labels were put into NumPy arrays and saved to \texttt{.npy} files for easy further loading and processing.

\subsection{Preprocessing CAMUS data}

We acquired the CAMUS data by going to their website (\url{http://camus.creatis.insa-lyon.fr/challenge/#challenges}). Once you get to their website, link on the first link, register on that website, and then you'll be free to download the dataset. 

In the CAMUS dataset, in addition to having view labels ('2CH' in their dataset is 'A2C' and likewise '4CH' is 'A4C'), they also label whether the view was taken in the end diastolic (ED) or end systolic (ES) portion of the cardiac cycle. We separated and took note of these labels, but we found no significant differences in the results.  

The raw CAMUS data came in .mhd format, a special file types used specifically for medical imaging. Through conversations with data creators, we discovered that the resolution for these images was lower in the x direction than the y direction and the way .mhd files compensate for a lower resolution is by adjusting the space between the pixels in that direction (indicated by the 'Element Spacing' field).
In order to convert to a standardized tiff file representation (where the spacing between pixels is uniform across width and height) we shrank the image in the y direction as:
\begin{align}
	y^* = \frac{y \cdot s_y}{s_x}
	\label{eq:camus_shrink_y}
\end{align}
where y is the original location (number of pixels) in the y direction, $s_y$ and $s_x$ are the spacing of pixels in the y and x directions (as given in the Element Spacing metadata), and $y^*$ is the new location the y direction. 

After this transformaion, the images were converted them to gray-scale, padded the shorter axis to achieve a square aspect ratio, and resized to 112 x 112 pixels. 

%Finally, all images and labels were put into numpy arrays and saved to .npy files.

\subsection{Further investigation of CAMUS performance}
In our main paper's Fig.~\ref{fig:results-heart2heart-view}, we assess how well our TMED-2-trained models, which get balanced accuracy in the range $92-96\%$ on TMED-2 test set, generalize to other external datasets.
The models did reasonably generalize to the Unity dataset (balanced accuracy ranges from $90-94\%$), however on the CAMUS dataset we saw all methods reach somewhat surprisingly lower overall performance (balanced accuracy $60-85\%$), though the \emph{relative} ranking of different methods was similar.

\begin{figure*}[!h]
\centering
\begin{tabular}{c c c}
\includegraphics[width=.3\textwidth]{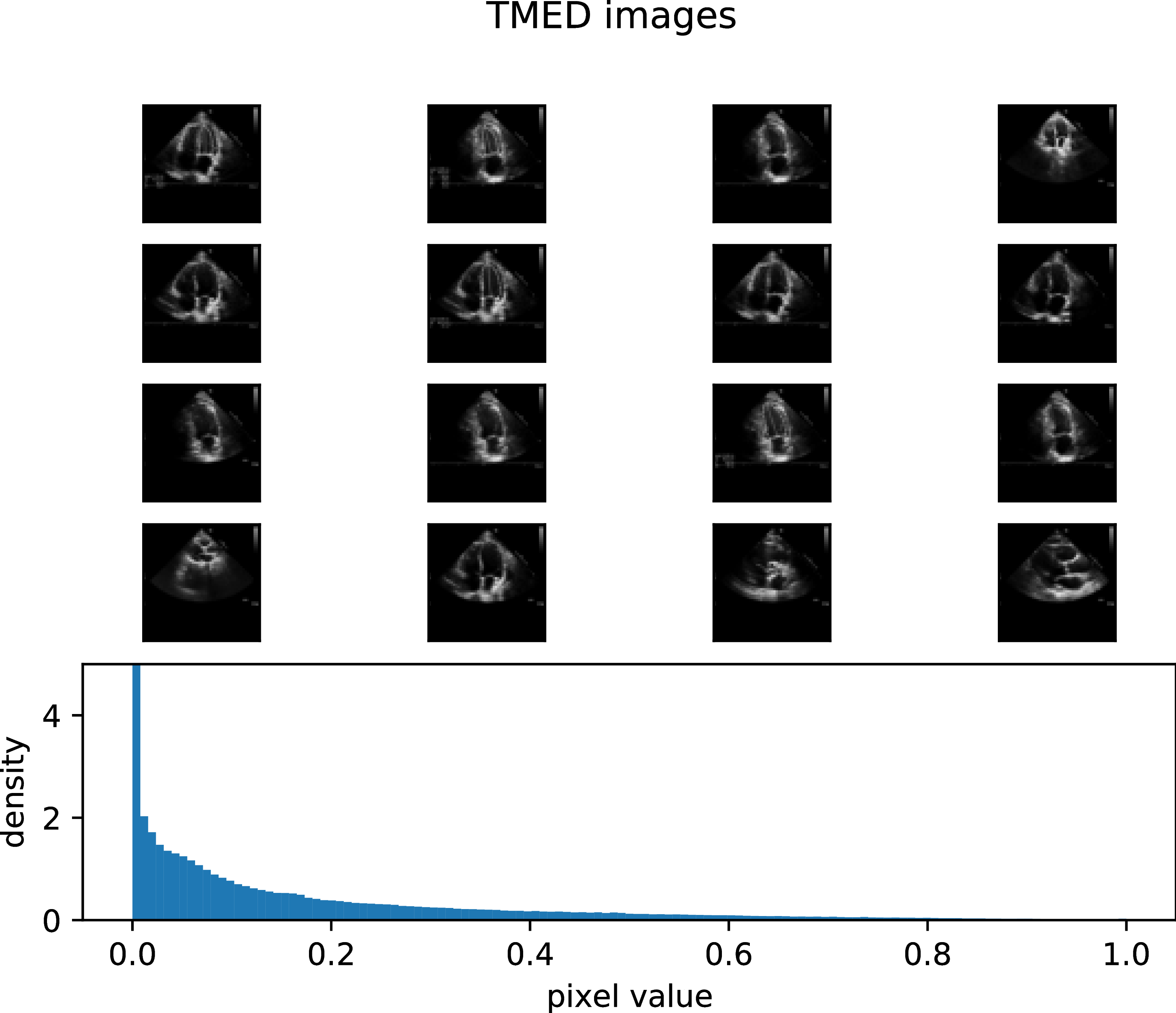}
&
\includegraphics[width=.3\textwidth]{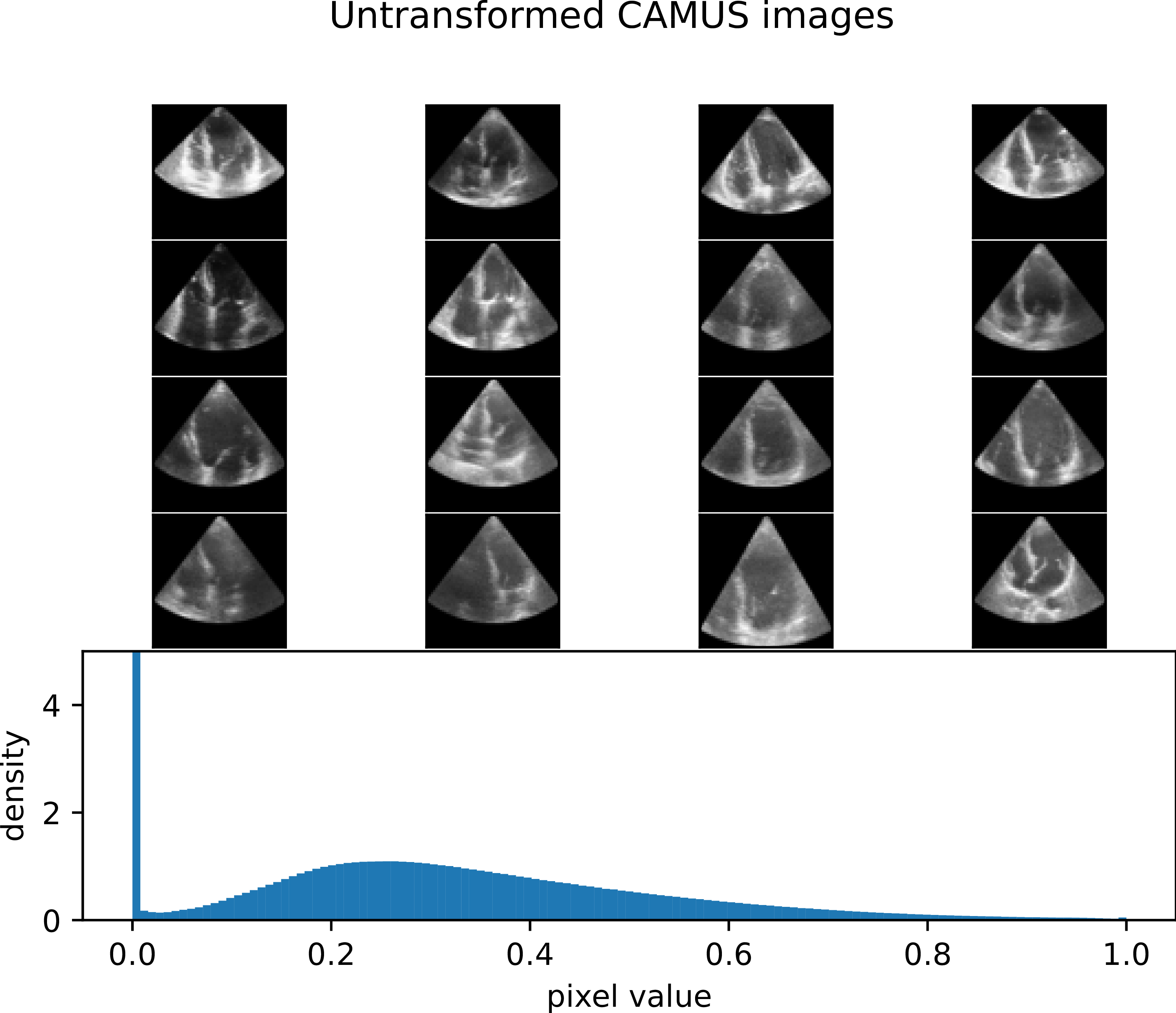}
&
\includegraphics[width=.3\textwidth]{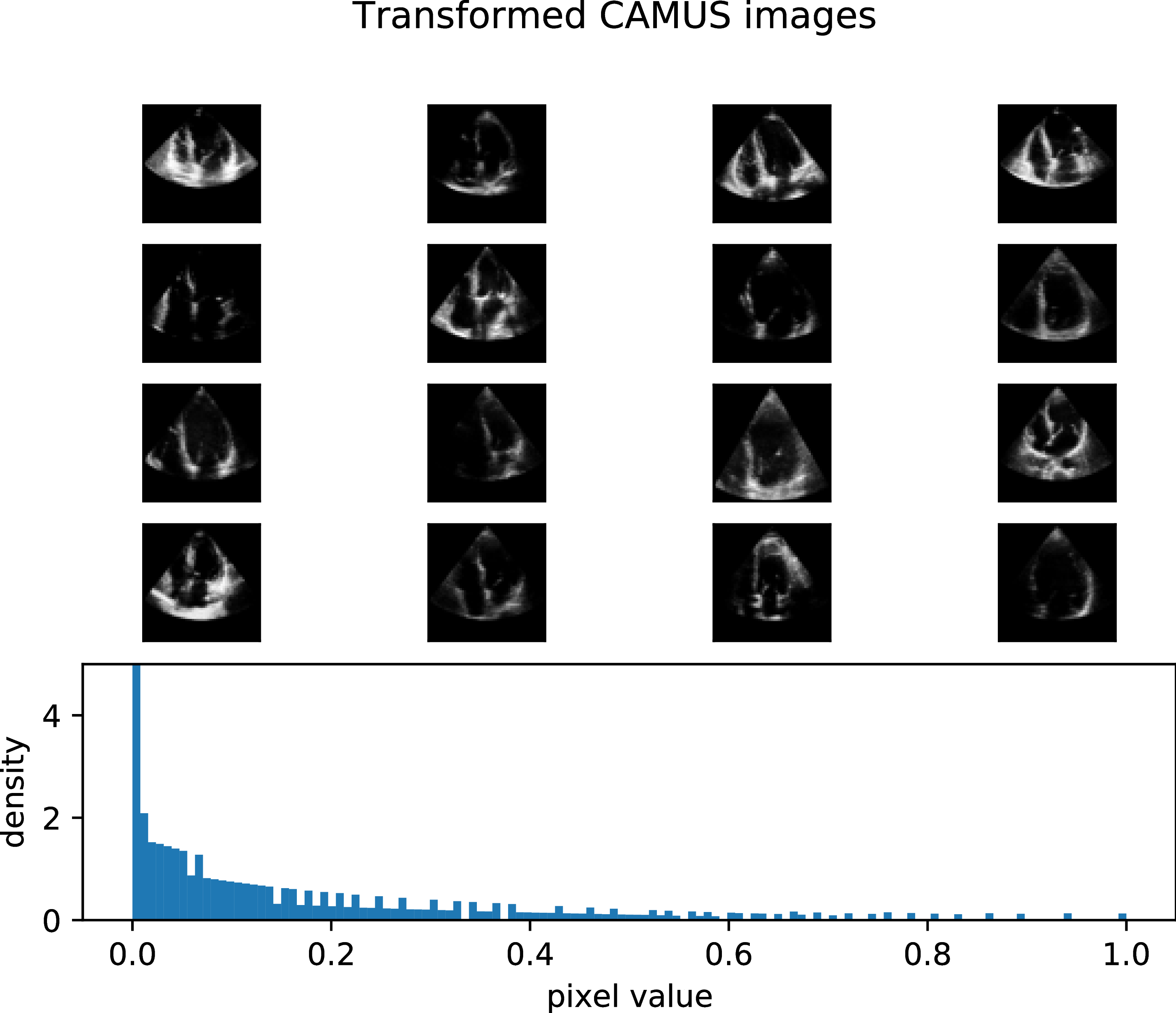}
\end{tabular}
\caption{
\textbf{A sample of images from the TMED-2 dataset, CAMUS dataset, and the same CAMUS pictures except under a pixel transformation to match the pixel intensity of TMED-2}
}%endcaption	
\label{fig:image_comparison}
\end{figure*}

\textbf{Visualizing differences.}
To investigate, we visually compared images from TMED-2, Unity, and CAMUS.
While Unity and TMED-2 looked similar, when comparing TMED-2 and CAMUS there are clear discrepancies in pixel intensity, likely from the use of a different ultrasound machine and different conventions standard intensity values and normalization.
Fig. ~\ref{fig:image_comparison} below provides sample images of the two datasets and a summary histogram of pixel intensity (aggregated across all images).

\textbf{Idea: Simple quantile transformation.}
To quickly try to remedy this discrepancy, we tried to transform the CAMUS images such that the pixel intensity distribution more closely resembles that of TMED-2.
In this transformation, we first mapped all the target pixels to its empirical quantile  (value between 0-1) and then we mapped that value to a pixel intensity in the source (TMED-2) images via the empirical inverse CDF. 
To see the effects of this transformation on the CAMUS images and on the pixel intensity histogram, look at the right-most panel of Fig.~\ref{fig:image_comparison}

\setlength{\tabcolsep}{.08cm}

\begin{figure*}[!h]
\centering
\begin{tabular}{c c}
\includegraphics[width=.36\textwidth]{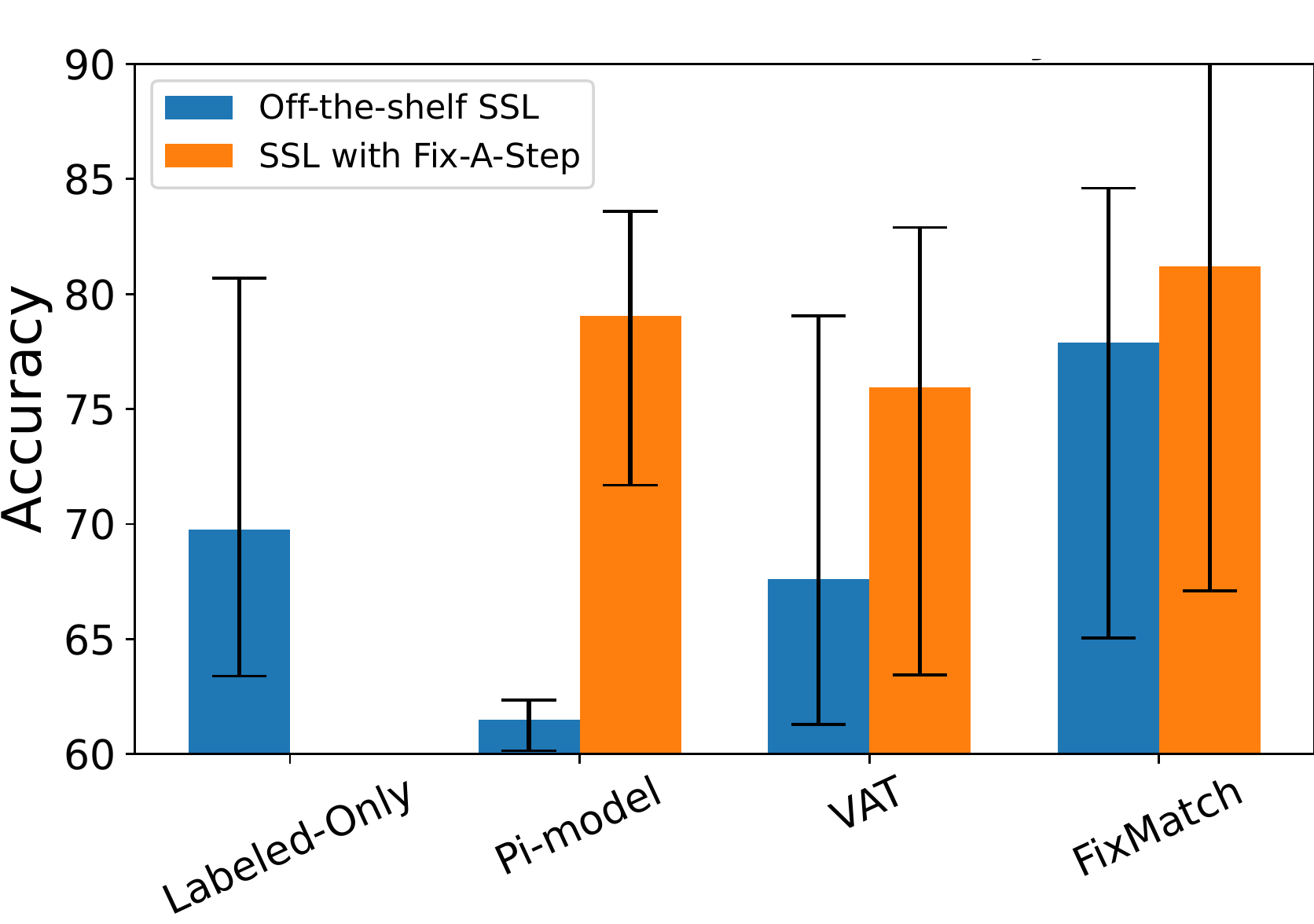}
&
\includegraphics[width=.36\textwidth]{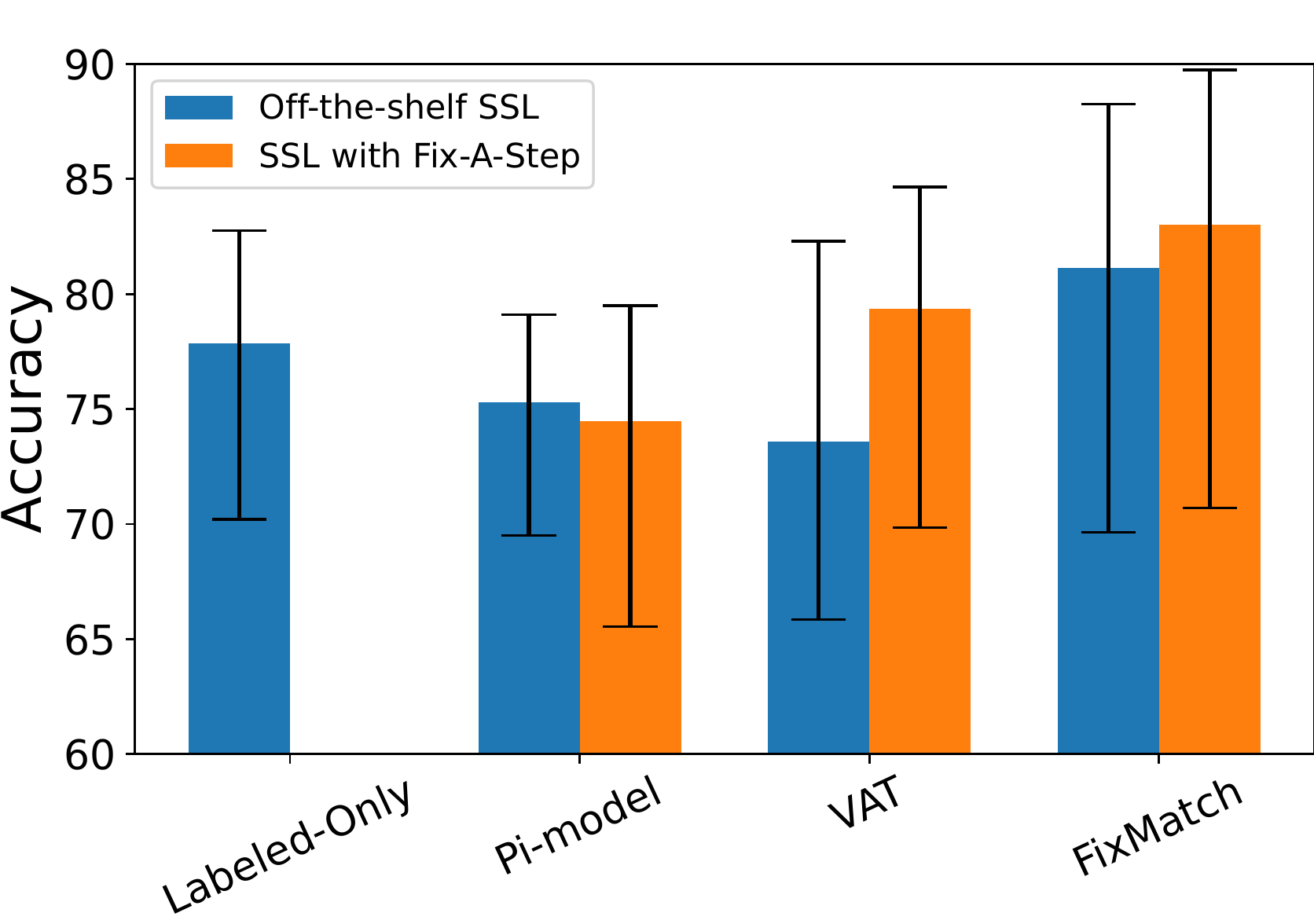}
\end{tabular}
\caption{
\textbf{Evaluation of the SSL methods from the paper on untransformed and transformed CAMUS images}
}%endcaption	
\label{fig:camus_transformation}
\end{figure*}

\vspace{1cm}
\textbf{Results after transform.}
The accuracy of the TMED-2-trained classifiers on both untransformed and transformed CAMUS data can be viewed in Fig.~\ref{fig:camus_transformation}.
Like we said in the main paper, Fix-a-Step clearly improves SSL models in classifying CAMUS view types for the untransformed data.
However, while the transformation itself seems to help model performance overall, Fix-a-Step doesn't seem to help as much in the transformed dataset (some gains for VAT, but both FixMatch and Pi-model the before-after difference seems negigible).
Importantly, Fix-A-Step is still \emph{competitive} with its base method, just not notably superior to it.
Much more work is needed here.
In the future, we hope to explore other ways to improve performance on the CAMUS dataset so it reaches accuracy levels seen in the Unity dataset. 

\textbf{Further investigation: Differences across splits.}
After investigating the results of the three data splits, we noticed that the first split seemed to significantly under perform on the CAMUS dataset, specifically with the A4C class. When we took a look at the Unity data for this split, we also noticed that, while the discrepancy wasn't as drastic, the A4C class did under perform when compared to the other classes. These results can be clearly seen in Tab. ~\ref{tab:a4c_class_accuracies}. In all method-dataset pairs, A4C performs significantly worse than other classes. 

A hypothesis we have is that this data split significantly under represents A4C and thus is not able to predict it as well. The reason why we don't see TMED-2 and Unity significantly under perform in this split in terms of total balance accuracy is because the other classes are a significant portion of their test sets so they're not as affected by A4C under performing; however, 50\% of CAMUS is A4C, so that dataset is affected to a higher degree. However, we are unsure as to why the A4C class accuracy in CAMUS does significantly worse than the A4C class accuracy in Unity. We will investigate this discrepancy further in the future. We think this open problem makes our Heart2Heart benchmark especially interesting.
\begin{table}[!h]
\centering
\begin{tabular}{c | c  c | c  c  c  }
\multicolumn{1}{c}{Methods} & \multicolumn{2}{c}{CAMUS} & \multicolumn{3}{c}{Unity}\\
 & A4C & A2C & A4C & A2C & PLAX 
 \\ \hline
Labeled-Only & \textbf{28.8} & 98.0 & \textbf{76.9} & 93.3 & 96.1\\
Pi-model & \textbf{27.1} &  96.8 & \textbf{81.9} & 93.1 & 98.2 \\
Pi-model w/ FAS & \textbf{45.4} & 98.0 & \textbf{87.5} & 94.1 & 99.6 \\
VAT & \textbf{26.6}& 98.3 & \textbf{77.0} & 93.1 & 97.7 \\
VAT w/ FAS  & \textbf{27.5}& 99.4& \textbf{84.6} & 96.2 & 95.5\\
Fix-Match & \textbf{34.1} & 96.0 & \textbf{79.8} & 94.8& 96.9\\
Fix-Match w/FAS & \textbf{36.3} & 97.9 & \textbf{83.5} & 95.2 & 99.0 \\

\end{tabular}
    \caption{
    \textbf{Class accuracies for data split 1 across methods for the Unity dataset and untransformed CAMUS dataset}. Bolded are the lowest class accuracies for each dataset-method pair.
    }%endcaption
    \label{tab:a4c_class_accuracies}
\end{table}

\section{METHODS SUPPLEMENT}
\label{app:methods}
Here, we provide implementation details of the two subprocedures in our Fix-A-Step training (Alg.~\ref{alg:FixAStep}).
Both procedures were originally proposed by MixMatch~\citep{berthelotMixMatchHolisticApproach2019}, we provide them here in common notation as the rest of our paper for clarity.

First, the algorithm \textsc{Aug+SoftLabel} is in Alg.~\ref{alg:aug_softlabel}. This procedure consumes a batch of raw images from the unlabeled set and returns two transformed batches, with a common set of ``sharpened'' soft (probabilistic) labels.

Second, the algorithm \textsc{MixMatchAug} is in Alg.~\ref{alg:mixmatch}. This procedure consumes a batch of raw labeled data, and produces a transformed batch of the same size.

\begin{algorithm}[!h]
\caption{Augment and Soft-Pseudo-Label}
\label{alg:aug_softlabel}
\textbf{Input}: Unlabeled batch features $\mathbf{x}^U$
\\
\textbf{Output}: Augmented features $\mathbf{x}^U_1, \mathbf{x}^U_2$, Soft pseudo labels $\tilde{\mathbf{y}}^U$
\\
\textbf{Hyperparameters}
\begin{itemize}
	\item Sharpening temperature $\tau{>}0$
\end{itemize}
\textbf{Procedure}
\begin{algorithmic}[1] %[1] enables line numbers
\FOR{each image $x$ in $\mathbf{x}^U$}
\STATE $x^{(1)} \gets \text{BasicImageAugment}(x_n)$
\STATE $x^{(2)} \gets \text{BasicImageAugment}(x_n)$
\STATE $\rho^{(1)} \gets f_w( x^{(1)} )$ \texttt{~~~~~~~~// Probability vector predicted by neural net}
\STATE $\rho^{(2)} \gets f_w( x^{(2)} )$
\STATE $\tilde{r} \gets \left( 
	  \frac{1}{2}\rho^{(1)} 
	+ \frac{1}{2}\rho^{(2)} \right)
	^{1 / \tau}$ \texttt{~~// Non-negative vector, sharpened by element-wise power}
\STATE $S \gets \sum_c \tilde{r}_c$
\STATE $\tilde{y} \gets [
	\frac{\tilde{r}_1}{S},
	\frac{\tilde{r}_2}{S},
	\ldots
	\frac{\tilde{r}_C}{S}]$ \texttt{~~~~~~// Normalize to ``soft'' label (proba. vector)}
\STATE Add $x^{(1)}$ to $\tilde{\mathbf{x}}^U_1$
\STATE Add $x^{(2)}$ to $\tilde{\mathbf{x}}^U_2$
\STATE Add $\tilde{y}$ to $\tilde{\mathbf{y}}^U$
\ENDFOR
\STATE \textbf{return} $\tilde{\mathbf{x}}^U_1$, $\tilde{\mathbf{x}}^U_2$, $\tilde{\mathbf{y}}^U$
\end{algorithmic}
\end{algorithm}

% \begin{algorithm}[tb]
% \caption{Fix-A-Step Training}
% \label{alg:FixAStep}
% \textbf{Input}: Labeled set $\mathcal{D}^L$, Unlabeled set $\mathcal{D}^U$ (uncurated)
% \\
% \textbf{Output}: Trained weights $w^*$
% \\
% \textbf{Hyperparameters}
% \begin{itemize}
% 	\item Shape parameter $\alpha{>}0$ of Beta dist. for {\small \textsc{MixMatchAug}}
% 	\item Initial weights $w$, Max. iterations $T$
% 	\item Unlabeled-loss weight per iter $\lambda_1, \ldots \lambda_T$
% \end{itemize}
% \begin{algorithmic}[1] %[1] enables line numbers
% \FOR{iter $t \in 1, 2, \ldots T$ until converged}
% \STATE $\{\mathbf{x}^L, \mathbf{y}^L\}, \mathbf{x}^U \gets {\small \textsc{GetNextMinibatch}}()$
% \STATE $\tilde{\mathbf{x}}^U, \tilde{\mathbf{y}}^U \gets {\small \textsc{Augment+SoftLabel}}(\mathbf{x}^U; w)$
% \STATE $\tilde{\mathbf{x}}^L, \tilde{\mathbf{y}}^L \gets 
% {\small \textsc{MixMatchAug}}(\{\mathbf{x}^L, \mathbf{y}^L\}, \tilde{\mathbf{x}}^U, \tilde{\mathbf{y}}^U; \alpha)$
% \STATE $g^L \gets \nabla_w \ell^L( \tilde{\mathbf{x}}^L, \tilde{\mathbf{y}}^L; w)$
% \STATE $g^U \gets \nabla_w \ell^U( \tilde{\mathbf{x}}^U, \tilde{\mathbf{y}}^U; w)$
% \IF {${\small \textsc{InnerProduct}}(g^L,g^U) > 0$}
% \STATE $w \gets w - \epsilon (g^L + \lambda_t g^U)$
% \ELSE
% \STATE $w \gets w - \epsilon g^L$
% \ENDIF
% \ENDFOR
% \STATE \textbf{return} w
% \end{algorithmic}
% \end{algorithm}

\begin{algorithm}[!h]
\caption{MixMatchAug : Transformation of Labeled Set}
\label{alg:mixmatch}
\textbf{Input}:
Labeled batch $\mathbf{x}^L, \mathbf{y}^L$, 
Unlabeled batch $\mathbf{x}^U, \tilde{\mathbf{y}}$, 
\\
\textbf{Output}: 
Transformed labeled batch $\mathbf{\tilde{x}}^L, \mathbf{\tilde{y}}^L$
\\
\textbf{Hyperparameters}
\begin{itemize}
	\item Shape $\alpha{>}0$ of $\text{Beta}(\alpha, \alpha)$ dist.
\end{itemize}
\begin{algorithmic}[1] %[1] enables line numbers
\FOR{image-label pair $x, y$ in labeled batch ${\mathbf{x}^L, \mathbf{y}^L}$}
\STATE $x', y' \gets \textsc{SampleOnePair}([\mathbf{x}^L, \tilde{\mathbf{x}}^U_1, \tilde{\mathbf{x}}^U_2],
     [\mathbf{y}^L, \tilde{\mathbf{y}}^U, \tilde{\mathbf{y}}^U])$
\STATE $\beta' \sim \textsc{SampleFromBeta}(\alpha, \alpha)$
\STATE $\beta \gets \textsc{Max}(\beta', 1-\beta')$ 
\STATE $\tilde{x} \gets \beta x + (1-\beta) x'$
\STATE $\tilde{y} \gets \beta y + (1-\beta) y'$
\STATE Add $\tilde{x}$ to $\tilde{\mathbf{x}}^L$
\STATE Add $\tilde{y}$ to $\tilde{\mathbf{y}}^L$
\ENDFOR
\STATE \textbf{return} $\tilde{\mathbf{x}}^L, \tilde{\mathbf{y}}^L$
\end{algorithmic}
\end{algorithm}

\textbf{Fix-A-Step with FixMatch}. 
Here, we further clarify how Fix-A-Step works with FixMatch-base. In the original FixMatch, each unlabeled image generates one weakly augmented version and one strongly-augmented. With
Fix-A-Step, an additional weakly augmented version is generated. The two weak images are used in the Fix-A-Step augmentation phase to transform the labeled set. The unlabeled loss is calculated using one weak and one strong image as in FixMatch. In this work, we used the original images for \textit{unlabeled loss} calculation (not the transformed images from FixAStep augmentation phase) so that only the unlabeled set affect the labeled set but not vise versa, since we focus on analyzing the value of the unlabeled set to the labeled set. Network parameters are updated via Algorithm~\ref{alg:FixAStep} line 5-7.

\section{RELATED WORK SUPPLEMENT}
\label{app:related_work}
\paragraph{SSL benchmarks.}
SSL methods continue to focus a few datasets intended for \emph{fully-supervised} image classification, such as SVHN, CIFAR-10, CIFAR-100, and ImageNet.
% MCH: not sure we should cite ~\citep{su2021realistic} here.
This is a problem because these data are post-hoc repurposed for SSL, dropping known labels to create unlabeled sets in \emph{artificial} fashion.
The resulting unlabeled sets are \emph{too curated}: images usually come from the same classes as the labeled set with similar frequencies.
However, real applications that motivate SSL require an easy-to-acquire unlabeled set that is \emph{uncurated}. 

Recent research has further identified problems with CIFAR and ImageNet.
First, 3\% of CIFAR-10 and 10\% of CIFAR-100 test images have perceptually-indistinguishable duplicates in the train set~\citep{barzWeTrainTest2020}.
This questions whether high-scoring methods are memorizing rather than truly generalizing.
Second, a notable fraction ($\sim$5\%) of the labels in the test sets of CIFAR-100 and ImageNet data are \emph{incorrect}~\citep{northcuttPervasiveLabelErrors2021}.
More generally, overuse of the same benchmarks over decades may lead to over-optimistic assessments of heldout error rates~\citep{yadavColdCaseLost2019} and may privilege methods that exploit shortcuts or biases in the available data that hurt true generalization~\citep{tsiprasImageNetImageClassification2020,geirhosShortcutLearningDeep2020}.
Given this background, we argue that new SSL benchmarks motivated by intended applications are sorely needed to help ensure the next-generation of SSL methods delivers on its promise of generalization.

\paragraph{Gradient step modifications.} Recently, across many sub-areas of ML that optimize of a multi-task loss, modifying the direction of gradient descent updates during training has born fruit.

The idea of gradient matching has been proposed to solve catastrophic forgetting problems in continual learning \citep{lopez-pazGradientEpisodicMemory2017, chaudhry2018efficient, riemer2018learning, zeng2019continual, farajtabar2020orthogonal}.
In \citet{lopez-pazGradientEpisodicMemory2017}, the author proposed a method called Gradient Episodic Memory (GEM), where they used a memory bank to store representative samples of previous tasks.
While minimizing the loss on current task, they use the inner product of the gradient between current and previous tasks as an inequality constraint.
In \citet{chaudhry2018efficient}, Averaged GEM (A-GEM) is proposed as an improved version of GEM.
A-GEM ensures that at every training step the average episodic memory loss over the previous tasks does not increase. 
\citet{riemer2018learning} formally proposed the transfer-interference trade-off perspective for looking at the application of gradient matching in continual learning, which defines whether helpful transfer or interference occurs between two labeled examples in terms of the inner product of gradients with respect to parameters evaluated at those examples.
%For two distinct examples $(x_i, y_i)$ and $(x_j, y_j)$, transfer happens when $\frac{\partial L(xi, yi)}{\partial \theta}  \frac{\partial L(xj,yj)}{\partial \theta} > 0$, given current network parameter $\theta$ and loss function $L$, and interference otherwise. 
\citet{zeng2019continual} developed Orthogonal Weights Modification (OWM) method to project the weight updates to the orthogonal direction to the subspace spanned by previously learned task inputs while 
\citet{farajtabar2020orthogonal} projects the new task's gradient to the direction that is perpendicular to the gradient space of previous tasks.

Similar ideas were later used in multi-task learning \citep{duAdaptingAuxiliaryLosses2020, yuGradientSurgeryMultiTask2020}, domain generalization \cite{shi2021gradient} and neural architecture search \citep{gong2021nasvit}.

%\paragraph{Deep SSL for medical images.}
%Applications of SSL to medical imaging ~\citep{madaniDeepEchocardiographyDataefficient2018,calderon-ramirezDealingScarceLabelled2021} are exciting but relatively rare.

\section{REPRODUCIBILITY SUPPLEMENT}
\label{app:experiment_details}
\label{app:hyperparameter_list}
\subsection{Codebase}
Our work builds upon several public repositories that represent either official or well-designed third-party implementations of popular SSL methods.

\begin{table}[!h]
\begin{tabular}{l l l}
Method & Code URL & notes
\\
FixMatch & \url{github.com/google-research/fixmatch}
         & original
\\
         & \url{github.com/kekmodel/FixMatch-pytorch} & PyTorch version
\\
MixMatch & \url{github.com/google-research/mixmatch}
		 & original
\\
         & \url{github.com/YU1ut/MixMatch-pytorch} & PyTorch version
\\
Realistic SSL Eval. & \url{github.com/perrying/realistic-ssl-evaluation-pytorch}
\end{tabular}
\caption{Code repositories that we built upon to perform our experiments and verify the quality of results.}	
\end{table}

\subsection{Hyperparameters for CIFAR-10/ CIFAR-100}

Table~\ref{tab:hyperparameters_cifar} lists the experimental settings (dataset sizes, etc.) and hyperparameters used for all CIFAR-10/CIFAR-100 baselines.
We emphasize that \textbf{we \emph{not} tune any hyperparameters specifically for Fix-A-Step}: whenever we combined a base model with Fix-A-Step (e.g. Mean Teacher + Fix-A-Step), we simply copied the relevant hyperparameters for the base model from Table~\ref{tab:hyperparameters_cifar}, and set Fix-A-Step's unique hyperparameters to defaults $\alpha = 0.5, \tau=0.5$.

\setlength{\tabcolsep}{.08cm}
\newcommand{\WWW}{5.5cm}
\newcommand{\RRR}{2cm}
\begin{table}[!h]
	\setlength\extrarowheight{-10pt}
    \begin{tabular}{c c}
		{\Large BASIC SETTINGS CIFAR-10}
		&
		{\Large BASIC SETTINGS CIFAR-100}
		\\
		\begin{minipage}[t]{.5\textwidth}
		\strut\vspace*{-\baselineskip}\newline
			\begin{tabular}{L{\WWW}  R{\RRR}}
\\ \hline
Train labeled set size & 2400/300
\\ \hline
Train unlabeled set size & 16400/17800
\\ \hline
Validation set size & 3000
\\ \hline
Test set size & 6000
\end{tabular}	
		\end{minipage}
		&
		\begin{minipage}[t]{.5\textwidth}
		\strut\vspace*{-\baselineskip}\newline
			\begin{tabular}{L{\WWW}  R{\RRR}}
\\ \hline
Train labeled set size & 5000
\\ \hline
Train unlabeled set size & 17500
\\ \hline
Validation set size & 2500
\\ \hline
Test set size & 5000
\end{tabular}	
		\end{minipage}
		\\ %%%%%%%%%%%%%%%%%%%%%%%%%%
		\\ %%%%%%%%%%%%%%%%%%%%%%%%%%
		{\Large Labeled only}
		&
		{\Large VAT}
		\\
		\begin{minipage}[t]{.5\textwidth}
		\strut\vspace*{-\baselineskip}\newline
			\begin{tabular}{L{\WWW}  R{\RRR}}
\\ \hline
Labeled batch size & 64
\\ \hline
Learning rate & 3e-3
\\ \hline
Weight decay & 2e-3
\end{tabular}	
		\end{minipage}
		&
		\begin{minipage}[t]{.5\textwidth}
		\strut\vspace*{-\baselineskip}\newline
			\begin{tabular}{L{\WWW}  R{\RRR}}
\\ \hline
Labeled batch size & 64
\\ \hline
Unlabeled batch size & 64
\\ \hline
Learning rate & 3e-2
\\ \hline
Weight decay & 4e-5
\\ \hline
Max consistency coefficient & 0.3
\\ \hline
Unlabeled loss warmup iterations & 419430
\\ \hline
Unlabeled loss warmup schedule & linear
\\ \hline
VAT $\xi$ & 1e-6
\\ \hline
VAT $\epsilon$ & 6
\end{tabular}	
		\end{minipage}
		\\ %%%%%%%%%%%%%%%%%%%%%%%%%%
		\\ %%%%%%%%%%%%%%%%%%%%%%%%%%
		{\Large Pseudo-label}
		&
		{\Large Mean Teacher}
		\\
		\begin{minipage}[t]{.5\textwidth}
		\strut\vspace*{-\baselineskip}\newline
			\begin{tabular}{L{\WWW}  R{\RRR}}
\\ \hline
Labeled batch size & 64
\\ \hline
Unlabeled batch size & 64
\\ \hline
Learning rate & 3e-2
\\ \hline
Weight decay & 5e-4
\\ \hline
Max consistency coefficient & 1.0
\\ \hline
Unlabeled loss warmup iterations & 419430
\\ \hline
Unlabeled loss warmup ischedule & linear
\\ \hline
Pseudo-label threshold & 0.95
\end{tabular}	
		\end{minipage}
		&
		\begin{minipage}[t]{.5\textwidth}
		\strut\vspace*{-\baselineskip}\newline
			\begin{tabular}{L{\WWW}  R{\RRR}}
\\ \hline
Labeled batch size & 64
\\ \hline
Unlabeled batch size & 64
\\ \hline
Learning rate & 3e-2
\\ \hline
Weight decay & 5e-4
\\ \hline
Max consistency coefficient & 50.0
\\ \hline
Unlabeled loss warmup iterations & 419430
\\ \hline
Unlabeled loss warmup schedule & linear
\end{tabular}	
		\end{minipage}
		\\ %%%%%%%%%%%%%%%%%%%%%%%%%%
		\\ %%%%%%%%%%%%%%%%%%%%%%%%%%
		{\Large Pi-Model}
		&
		{\Large MixMatch}
		\\
		\begin{minipage}[t]{.5\textwidth}
		\strut\vspace*{-\baselineskip}\newline
			\begin{tabular}{L{\WWW}  R{\RRR}}
\\ \hline
Labeled batch size & 64
\\ \hline
Unlabeled batch size & 64
\\ \hline
Learning rate & 3e-2
\\ \hline
Weight decay & 5e-4
\\ \hline
Max consistency coefficient & 10.0
\\ \hline
Unlabeled loss warmup iterations & 419430
\\ \hline
Unlabeled loss warmup schedule & linear
\end{tabular}	
		\end{minipage}
		&
		\begin{minipage}[t]{.5\textwidth}
		\strut\vspace*{-\baselineskip}\newline
			\begin{tabular}{L{\WWW}  R{\RRR}}
\\ \hline
Labeled batch size & 64
\\ \hline
Unlabeled batch size & 64
\\ \hline
Learning rate & 3e-2
\\ \hline
Weight decay & 4e-5
\\ \hline
Max consistency coefficient & 75.0
\\ \hline
Unlabeled loss warmup iterations & 1048576
\\ \hline
Unlabeled loss warmup schedule & linear
\\ \hline
Sharpening temperature & 0.5
\\ \hline
Beta shape $\alpha$ & 0.75
\end{tabular}	
		\end{minipage}
		\\ %%%%%%%%%%%%%%%%%%%%%%%%%%
		\\ %%%%%%%%%%%%%%%%%%%%%%%%%%
		{\Large FixMatch}
		&
		{\Large OpenMatch}
		\\
		\begin{minipage}[t]{.5\textwidth}
		\strut\vspace*{-\baselineskip}\newline
			\begin{tabular}{L{\WWW}  R{\RRR}}
\\ \hline
Labeled batch size & 64
\\ \hline
Unlabeled batch size & 448
\\ \hline
Learning rate & 3e-2
\\ \hline
Weight decay & 5e-4
\\ \hline
Max consistency coefficient & 1.0
\\ \hline
Unlabeled loss warmup iterations & No warmup
\\ \hline
Unlabeled loss warmup schedule & No warmup
\\ \hline
Sharpening temperature & 1.0
\\ \hline
Pseudo-label threshold & 0.95
%\\ \hline
%Unlabeled/Labeled batch size ratio $\mu$ & 7
\end{tabular}	
		\end{minipage}
		&
		\begin{minipage}[t]{.5\textwidth}
		\strut\vspace*{-\baselineskip}\newline
			\begin{tabular}{L{\WWW}  R{\RRR}}
\\ \hline
Labeled batch size & 64
\\ \hline
Unlabeled batch size & 128
\\ \hline
Learning rate & 0.03
\\ \hline
Weight decay & 5e-4
\\ \hline
Lambda socr & 0.5
\\ \hline
Lambda oem & 0.1
\\ \hline
Warmup epoch before FixMatch & 10
\\ \hline
Unlabeled loss warmup iterations & No warmup
\\ \hline
Unlabeled loss warmup schedule & No warmup
\\ \hline
Sharpening temperature & 1.0
\\ \hline
Pseudo-label threshold & 0.0
%\\ \hline
%Unlabeled/Labeled batch size ratio $\mu$ & 7
\end{tabular}	
		\end{minipage}
		\\ %%%%%%%%%%%%%%%%%%%%%%%%%%
		\\ %%%%%%%%%%%%%%%%%%%%%%%%%%
		{\Large MTCF}
		&
		{\Large DS3L}
		\\
		\begin{minipage}[t]{.5\textwidth}
		\strut\vspace*{-\baselineskip}\newline
			\begin{tabular}{L{\WWW}  R{\RRR}}
\\ \hline
Domain batch size & 64
\\ \hline
Labeled batch size & 64
\\ \hline
Unlabeled batch size & 64
\\ \hline
Learning rate & 3e-4
\\ \hline
Weight decay & 6e-6
\\ \hline
Max consistency coefficient & 75
\\ \hline
Warmup epochs & 100
\\ \hline
Sharpening temperature & 0.5
\\ \hline
Beta shape $\alpha$ & 0.75
%\\ \hline
%Unlabeled/Labeled batch size ratio $\mu$ & 7
\end{tabular}	
		\end{minipage}
		&
		\begin{minipage}[t]{.5\textwidth}
		\strut\vspace*{-\baselineskip}\newline
			\begin{tabular}{L{\WWW}  R{\RRR}}
\\ \hline
Labeled batch size & 64
\\ \hline
Unlabeled batch size & 64
\\ \hline
Learning rate & 3e-4
\\ \hline
learning rate meta & 0.001
\\ \hline
learning rate wnet & 6e-5
\\ \hline
Max consistency coefficient & 10.0
\\ \hline
Unlabeled loss warmup iterations & 200000
\\ \hline
Unlabeled loss warmup schedule & sigmoid

%\\ \hline
%Unlabeled/Labeled batch size ratio $\mu$ & 7
\end{tabular}	
		\end{minipage}
		\\ %%%%%%%%%%%%%%%%%%%%%%%%%%
		\\ %%%%%%%%%%%%%%%%%%%%%%%%%%

	\end{tabular}
	\vspace{-.4cm}
\caption{
\textbf{Hyperparameters used for CIFAR experiments.}
All settings represent the recommended defaults suggested in implementations by original authors for the 400 examples/class setting.
We did \emph{not} tune any hyperparameters specifically for Fix-A-Step.
}%endcaption
\label{tab:hyperparameters_cifar}
\end{table}

\clearpage
\subsection{Hyperparameters for Heart2Heart}
As in all other experiments, hyper-parameters were not tuned at all for Fix-A-Step in our Heart2Heart evaluations.
Instead, to ensure fair comparisons (and in fact to make Fix-A-Step prove that its worth comes from something other than hyperparameter tuning), we did allow tuning hyperparameters for \emph{all other} methods except Fix-A-Step: the labeled-set-only baseline, the Open-Match baseline, and the basic off-the-shelf SSL methods Pi-model, VAT and FixMatch. 

For those methods that were allowed tuning, we ran 100 trials\footnote{in practice, for each trial we train for only 180 epochs to speed up the hyper-parameters selection process} of Tree-structured Parzen Estimator (TPE) based black box optimization using an open source AutoML toolkit\footnote{https://github.com/microsoft/nni} for each algorithm and each data split. The chosen hyper-parameters are then directly applied to Fix-A-Step without retuning. After hyper-parameter selection, each algorithm is then trained for 1000 epochs, the balanced test accuracy at maximum validation balanced accuracy is then reported.

\paragraph{Labeled-only:} we search learning rate in $\{0.001, 0.003, 0.01, 0.03, 0.1, 0.3\}$, weight decay in $\{0.0, 0.00005, 0.0005, 0.005, 0.05\}$, optimizer in $\{\text{Adam}, \text{SGD}\}$, learning rate schedule in $\{\text{Fixed}, \text{Cosine}\}$. Batch size is set to 64.

\paragraph{Pi-model:} 
We search learning rate in $\{0.003, 0.01, 0.03, 0.1\}$, weight decay in $\{0.0, 0.0005, 0.005, 0.05\}$, optimizer in $\{\text{Adam}, \text{SGD}\}$, 
learning rate schedule in $\{\text{Fixed}, \text{Cosine}\}$, Max consistency coefficient in $\{1.0, 5.0, 10.0, 20.0, 100.0\}$, unlabeled loss warmup iterations in $\{0, 17000, 34000\}$. Labeled batch size is set to 64 and unlabeled batch size is set to 64.

\paragraph{VAT:} 
We search learning rate in $\{0.0002, 0.0006, 0.002, 0.006\}$, weight decay in $\{0.000004, 0.00004, 0.0004\}$, optimizer in $\{\text{Adam}, \text{SGD}\}$, 
learning rate schedule in $\{\text{Fixed}, \text{Cosine}\}$, Max consistency coefficient in $\{0.3, 0.1, 0.9, 0.03, 3\}$,  unlabeled loss warmup iterations in $\{0, 17000, 34000\}$. Labeled batch size is set to 64, unlabeled batch size is set to 64. $\xi$ is set to 0.000001 and $\epsilon$ is set to 6.

\paragraph{FixMatch:} 
We search learning rate in $\{0.003, 0.01, 0.03, 0.1\}$, weight decay in $\{0.0005, 0.005, 0.05\}$, optimizer in $\{\text{Adam}, \text{SGD}\}$, 
learning rate schedule in $\{\text{Fixed}, \text{Cosine}\}$, Max consistency coefficient in $\{0.5, 1.0, 5.0, 10.0\}$, Labeled batch size is set to 64, unlabeled batch size is set to 320. We set sharpening temperature to 1.0 and pseudo-label threshold is set to 0.95 (as in CIFAR experiments).

\paragraph{Open-Match:} 
We search learning rate in $\{0.003, 0.01, 0.03, 0.1, 0.3\}$, weight decay in $\{0.0000005, 0.000005, 0.00005, 0.0005, 0.005, 0.05\}$, 
lambda oem in $\{0.03, 0.1, 0.3, 1.0\}$, lambda socr in $\{0.25, 0.5, 1.0, 2.0\}$ (see OpenMatch paper for hyperparameter definitions). Labeled batch size is set to 64, unlabeled batch size is set to 128, and all other hyperparameters following the author's released code.

\subsection{Labeled loss implementation: Weighted cross entropy}

On many realistic SSL classification tasks, even the labeled set will have noticeably \emph{imbalanced} class frequencies.
For example, in the TMED-2 view labels, the four view types (PLAX, PSAX, A4C, A2C) differ in the number of available examples, with the rarest class (A2C) roughly 3x less common than the most common class (PLAX).
To counteract the effect of class imbalance, we use weighted cross-entropy for labeled loss, following prior works \citep{huangNewSemisupervisedLearning2021,wu2021reducing}. 
Let integer $c \in \{1, 2, \ldots C\}$ index the classes in the labeled set, and let $N_c$ denote the  number of images for class $c$.
Then when we compute the labeled loss $\ell^L$, we assign a weight $\omega_c > 0$ to the true class $c$ that is inversely proportional to the number of images $N_c$ of the class in the training set:
\begin{align}
\ell^L(x, c; w) = - \omega_c \log f_w(x)[c],
\qquad 
\omega_c = \frac{\prod_{k\neq c}{N_{k}}}{\sum_{j=1}^C\prod_{k \neq j}{N_{k}}}
\impliedby
\omega_c \propto \frac{1}{N_c}
\end{align}
Here $c$ denotes the integer index of the true class corresponding to image $x$, $w$ denotes the neural network weight parameters, and $f_w(x)[c]$ denotes the $c$-th entry of the softmax output vector produced by the neural network classifier.

\subsection{Cosine-annealing of learning rate.}
We found that several baselines were notably improved using the cosine-annealing schedule of learning rate suggested by \citep{sohnFixMatchSimplifyingSemisupervised2020}.
Cosine-annealing sets the learning rate at iteration $i$ to $\eta cos(\frac{7\pi i}{16 I})$, where $\eta$ is the initial learning rate, and $I$ is the total iterations.

To be extra careful, we tried to allow all open-set/safe SSL baselines to also benefit from cosine annealing.
\begin{itemize}
\item MTCF is trained using Adam following the author's implementation \citep{yuMultiTaskCurriculumFramework2020}.
Although the author did not originally use cosine learning rate schedule, we found that adding cosine learning rate schedule substantially improve MTCF's performance.
We thus report the performance for MTCF \emph{with cosine annealing}. 

\item DS3L is trained using Adam following the author's implementation \citep{guoSafeDeepSemiSupervised2020}.
We tried to add Cosine learning rate to DS3L, but this results in worse performance.
We thus report the performance for DS3L without cosine learning rate.

\end{itemize}

% \section{CIFAR-10 Class Mismatch description}
% \label{app:CIFAR10_Class_Mismatch_description}
% \input{app_CIFAR10_Class_Mismatch_description.tex}

\end{document}